\documentclass[11pt]{article}

\usepackage[final]{acl}

\usepackage{times}
\usepackage{latexsym}
\usepackage[T1]{fontenc}
\usepackage[utf8]{inputenc}
\usepackage{microtype}
\usepackage{inconsolata}
\usepackage{amsmath, amssymb, amsthm}
\usepackage{bm}
\usepackage{graphicx}
\usepackage{booktabs}
\usepackage{multirow}
\usepackage{makecell}
\usepackage{siunitx}
\usepackage{float}
\usepackage{dblfloatfix}   % fixes figure*[t!] placement in two-column mode
\usepackage{placeins}
\usepackage[dvipsnames, svgnames, x11names]{xcolor}
\usepackage{colortbl}
\usepackage{algorithm}
\usepackage{algpseudocode}
\usepackage{enumitem}
\usepackage{tikz}
\usetikzlibrary{arrows,arrows.meta,positioning,calc,decorations.pathreplacing}
\usepackage{pgfplots}
\pgfplotsset{compat=1.18}
\usepackage{cancel}
\definecolor{mist}{RGB}{230,220,245}    % highlight: GIST row (w/ RL block)
\definecolor{rlbg}{RGB}{245,243,250}    % highlight: other w/ RL block rows
\definecolor{qualityQwen}{HTML}{C5DCF5}
\definecolor{qualityTie}{HTML}{EDE9F5}
\definecolor{qualityGPT}{HTML}{D8C8EC}

\usepackage{pifont}
\newcommand{\cmark}{\ding{51}}  % ✓
\newcommand{\xmark}{\ding{55}}  % ✗
\newcommand{\codi}{CoDI}
\newcommand{\gist}{\codi}  % legacy alias; all new code should use \codi
\newcommand{\sg}{\textnormal{\textsc{SG}}}
\newcommand{\as}{\textnormal{\textsc{AS}}}
\newcommand{\pas}{p_{\mathrm{AS}}}       % Answer Solver probability shorthand
\newcommand{\dg}[1]{{\scriptsize\textcolor{NavyBlue}{$#1$\%}}}       % blue Δ (loss)
\newcommand{\dr}[1]{{\scriptsize\textcolor{BrickRed}{$+#1$\%}}}  % red Δ (gain)

\title{Cognitively Decoupled Meta-Induction for Out-of-Distribution Generalization}

\author{
\textbf{Yuqian Wang}$^{1}$ \quad
\textbf{Yuanzheng Wang}$^{2}$ \quad
\textbf{Chaojun Xiao}$^{3,\dagger}$ \quad
\textbf{Ke Xu}$^{4}$ \quad
\textbf{Bingxiang He}$^{3}$ \\
\textbf{Jinghan Tan}$^{5}$ \quad
\textbf{Zijun Chen}$^{6}$ \quad
\textbf{Lu Chen}$^{3,\dagger}$ \quad
\textbf{Yi-Xin Huo}$^{1,7}$ \quad
\textbf{Maosong Sun}$^{3,\dagger}$ \\
{\normalfont\small $^{1}$Beijing Institute of Technology \quad
$^{2}$University of Chinese Academy of Sciences} \\
{\normalfont\small $^{3}$Department of Computer Science and Technology, Tsinghua University} \\
{\normalfont\small $^{4}$Peking University \quad
$^{5}$Beijing University of Aeronautics and Astronautics} \\
{\normalfont\small $^{6}$Renmin University of China \quad
$^{7}$Muyuan Laboratory} \\
{\normalfont\footnotesize \texttt{3220252125@bit.edu.cn, xcjthu@gmail.com, chenlu6v@gmail.com, sms@tsinghua.edu.cn}}
}

\begin{document}
\maketitle
\begingroup
\renewcommand{\thefootnote}{\fnsymbol{footnote}}
\footnotetext[2]{\hspace{0.35em}Corresponding Authors.}
\endgroup

% ============================================================
\begin{abstract}
In-context learning (ICL) tackles downstream tasks conditioned on demonstrations.
This paradigm relies heavily on neighbor-based matching between queries and demonstrations, limiting out-of-distribution (OOD) performance.
For better generalization, rule-based inductive reasoning derives task-level regularities, which we term task-level inductive representations (TIR), before directly answering.
% To decouple these stages, we  
% Consequently, TIR-based inductive reasoning induces TIR .
However, existing methods lack generalizable inductive ability across diverse tasks.
% this requires meta-induction, the cross-task ability to infer query-independent TIR from supports, which standard ICL largely lacks.
Inspired by meta-learning, we propose Cognitive-Decoupling meta-Induction (CoDI) for TIR-based inductive reasoning on general tasks. CoDI comprises a \textit{Cognitive-Decoupling Framework} for query-blind TIR induction, and a \textit{Dual-Loop Meta-Inductive Learning} method for cross-task inducer optimization.
Empirically, CoDI achieves substantial performance gains on five ID and OOD benchmarks, and
preserves OOD gains without explicit induction at inference.
Code is available at \url{https://github.com/Yuqian-Wang2277/codi-meta-induction}.
\end{abstract}
% ============================================================
\section{Introduction}
\label{sec:intro}

\begin{figure}[t!]
\centering
\includegraphics[width=\columnwidth]{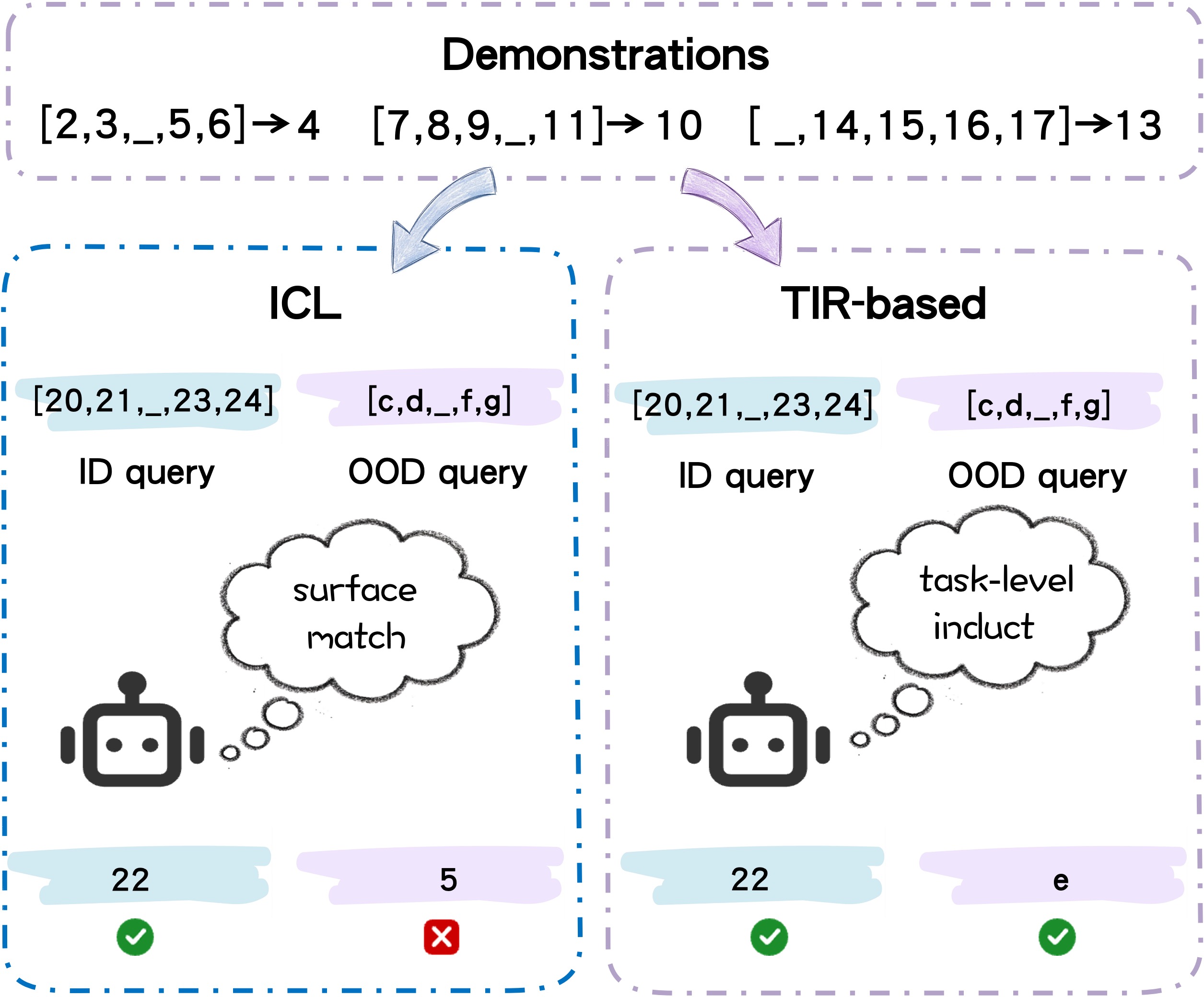}
\caption{%
  ICL (left) surface-matches and fails OOD.
  TIR-based reasoning (right) induces a task-level rule that generalizes to both ID and OOD. 
}
\label{fig:overview}
\end{figure}

% ── Para 1: 动机 + 背景介绍 ──
\noindent In-context learning (ICL) is a prevalent paradigm for tackling downstream tasks, allowing large language models (LLMs) to solve new queries conditioned on limited contextual labeled demonstrations~\citep{NEURIPS2020_1457c0d6}.
Despite achieving promising performance, ICL can rely heavily on demonstration format, label space, and nearest-neighbor-style similarity between queries and demonstrations~\citep{min2022rethinking,li2025mirage}.
Under distribution shifts and unseen task functions, such similarity-dominated behavior often suffers substantial performance degradation~\citep{wang2024incontextlearningreallygeneralize}.

% ── Para 2: 现有方法均无法产生 TIR ──
% \textbf{CoT与提示优化}虽拓展了少样本推理，却\textbf{难稳定形成}由支持集单独推导、与测试查询解耦、且可跨分布复用的\textbf{任务级归纳表征}（\textbf{TIR}）。
To address this limitation, rule-based inductive reasoning derives generalizable principles for straight answering, which we name task-level inductive representations (TIR).
The difference between ICL and TIR-based reasoning is shown in Figure~\ref{fig:overview}.
The former relies on surface matching with certain demonstrations, while the latter learns universal task patterns from all supports for better transfer.
% , which couples induction with query-specific straight answering, with CoDI's decoupled induction-before-straight-answering pipeline.
% abstract generalizable principles through inductive reasoning 
% ICL largely performs surface matching by aligning each query with the most similar support input and reusing its output pattern, so reasoning remains tied to particular demonstrations.
% TIR-based method instead induces task-level regularities verified across all $K$ supports and transferable to unseen queries within the task family.
Some works elicit intermediate reasoning through chain-of-thought (CoT) prompting and structured reasoning plans~\citep{wei2022cot,zhou2024selfdiscover}.
Yet these reasoning structures are usually produced at inference time for the prompted query or task context, so they are not learned as a reusable cross-task inducer~\citep{li2025mirage,zhou2024selfdiscover}.
A complementary direction pursues task-level rules via automated prompt
optimization~\citep{zhou2023ape,fernando2024promptbreeder,agrawal2026gepareflectivepromptevolution},
but such methods optimize prompts per task or system and must re-optimize when the task setting changes.

% 为解决这一问题，基于规则的方法通过归纳推理抽象出通用的规则，再运用于具体查询。部分工作通过思维链与结构化推理来增加这一中间步骤~\citep{wei2022cot,zhou2024selfdiscover}。然而，示例侧的规律抽取与查询侧预测同路耦合，容易生成\textbf{query-conditioned} 的解题思路，难以跨样本复用~\citep{li2025mirage,zhou2024selfdiscover}。另有工作通过自动化提示优化获取任务级通用规则~\citep{zhou2023ape,fernando2024promptbreeder,agrawal2025gepa}。然而它们多在单个任务上搜索任务特定的规则，每当跨任务运用时需要重新搜索，低效耗时。

% 优化信号缺少跨任务巩固归纳的机制，跨任务可迁移性受限。

% 综上，上述路径\textbf{均未稳定给出}与 \textbf{TIR} 判据一致的\textbf{任务级归纳表征}。

% 但仍以具体查询为条件，主体仍是

% Mirage 等评测揭示了 \textbf{neighbor-based} 与 \textbf{rule-based} 的不对称~\citep{li2025mirage}：在标准 ICL 中，归纳信号更易沿“示例—查询”表面对齐路径流动，而难以被固定为可脱离具体查询措辞、可反复调用的规则级表述。

% ── Para 3: TIR 可学习与查询盲约束 + 训练难点 ──
In comparison, human concept learning can infer compact, program-like rules from very few examples~\citep{lake2015human}, suggesting an induction-before-straight-answering pattern even on unfamiliar tasks.
We refer to this cross-task TIR inference capability as meta-induction.
Recent work finds that LLMs are weak at this capability~\citep{li2025mirage}, as their inductive behavior is dominated by neighbor-based matching rather than reusable abstract rules.
However, instilling this capacity in LLMs presents two challenges:
i)~ensuring meta-induction generalizes across tasks without overfitting to instance-specific patterns or task-local rules;
ii)~learning TIR in the absence of ground-truth labels and established evaluation criteria.
% because they learn to induce during their pass experience.
% because they first distill a reusable solution method from several similar demonstrations and then apply it to the target problem instead of copying a single demonstration.

% 相比之下，人类往往能熟练运用这种先归纳-后执行的推理范式~\citep{lake2015human}，即使是面对陌生任务。本文将这种面对不同任务都能基于示例归纳出任务级规律的能力称为\textbf{元归纳}（meta-induction）能力，并将\textbf{元归纳}面向特定任务时的显式输出形式化为\emph{任务级归纳表征}（task-level inductive representation, TIR）。Mirage评测揭示了常见LLM元归纳能力的不足，其归纳行为多为对特定输入输出的记忆与匹配，而非形成可反复调用的规则级表述~\citep{li2025mirage}。然而，对于元归纳能力的培养面临两大挑战：（1）如何保证元归纳能力的泛化性，避免对特定样例或特定任务的局部规则进行过度拟合？（2）如何指导TIR的生成，在TIR缺乏标注和衡量标准的情况下？

\begin{figure*}[t!]
\centering
\includegraphics[width=\textwidth]{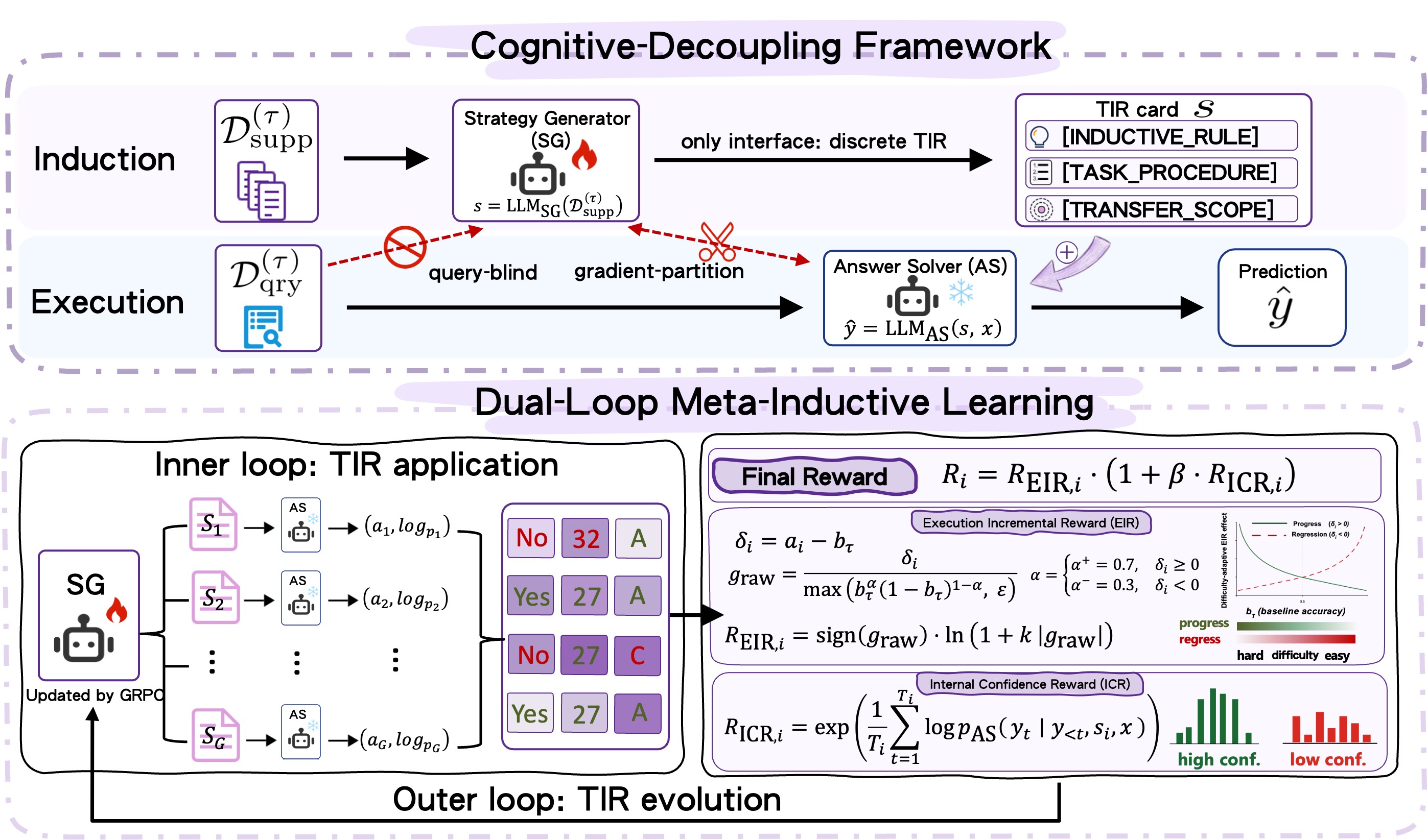}
\caption{%
  Overview of \codi{}.
  The \textbf{Cognitive-Decoupling Framework} induces query-blind TIR with \sg{} and executes each query with \as{}.
  \textbf{Dual-Loop Meta-Inductive Learning} applies candidate TIRs in the \textbf{inner loop} and optimizes \sg{} across tasks in the \textbf{outer loop} via GRPO with EIR and ICR.
}
\label{fig:arch}
\end{figure*}

% ── Para 4: CoDI framework and training signal ──
To address these challenges, we propose \textbf{Co}gnitively \textbf{D}ecoupled meta-\textbf{I}nduction (\textbf{CoDI}), a TIR-based inductive reasoning method on general tasks. 
As shown in Figure~\ref{fig:arch}, CoDI comprises a \textit{Cognitive-Decoupling Framework} and \textit{Dual-Loop Meta-Inductive Learning} method.
In the Cognitive-Decoupling Framework, a query-blind inducer induces TIR solely from demonstrations, and an executor executes TIR for each query.
Based on the above framework, Dual-Loop Meta-Inductive Learning optimizes the inducer across tasks via meta-RL, guided by execution feedback without ground-truth TIR labels, where an inner loop applies candidate TIR and an outer loop refines induction via two complementary rewards for extrinsic correctness and intrinsic certainty.

% Table~\ref{tab:qualitative_cases} contrasts these behaviors on matched OOD episodes.

% ── Para 5: main results ──
Experiments on BBH~\citep{suzgun2022bbh}, BIG-Bench~\citep{srivastava2023beyond}, MATH-500~\citep{lightman2023lets}, StrategyQA~\citep{geva2021strategyqa}, and ReClor~\citep{yu2020reclor} show that \codi{} (Qwen3-4B) outperforms ICL on in-distribution (ID) and out-of-distribution (OOD) evaluations.
On BIG-Bench, \codi{} achieves $+120\%$ and $+18\%$ gains on ID and OOD splits, respectively.
Furthermore, a smaller trained inducer surpasses a larger untrained one on OOD tasks, confirming TIR quality as an independently optimizable capability orthogonal to model scale.
Notably, OOD gains persist even without the induction step at inference, indicating that meta-inductive training internalizes inductive priors into model parameters.

In summary, our contributions are as follows:
\begin{itemize}[leftmargin=*, itemsep=2pt]
\item We formulate TIR-based inductive reasoning as a learning problem that requires inferring reusable task rules from support examples alone before execution.
\item We present \codi{} to instill meta-induction without TIR supervision, decoupling rule induction from straight answering and optimizing inducers across tasks via reinforcement learning on execution feedback.
\item Experiments on five benchmarks show \codi{} outperforming few-shot ICL in- and out-of-distribution, with smaller trained inducers beating larger untrained ones and OOD gains persisting without inference-time induction.
\end{itemize}
% ============================================================
\section{Problem Statement}
\label{sec:preliminaries}

\noindent In this section, we formalize the task of TIR-based inductive reasoning.

\paragraph{Notation.}
Let $\mathcal{T}$ denote a task space.
Each task $\tau \sim P(\mathcal{T})$ has a $K$-shot demonstration set
$\mathcal{D}_\text{supp}^{(\tau)} = \{(x_i, y_i)\}_{i=1}^{K}$
and query set $\mathcal{D}_\text{qry}^{(\tau)}$ of pairs $(x, y^*)$.
The TIR of $\tau$ is a natural-language strategy $s$ induced from $\mathcal{D}_\text{supp}^{(\tau)}$, and $\pi_{\mathrm{ind}}$ denotes the inducer policy, where $\mathrm{ind}$ abbreviates \emph{inducer}.

\paragraph{Task.}
Given $\mathcal{D}_\text{supp}^{(\tau)}$ and query $x$, TIR-based inductive reasoning samples
$s \sim \pi_{\mathrm{ind}}(\cdot \mid \mathcal{D}_\text{supp}^{(\tau)})$ as the TIR and uses an executor to predict
$\hat{y}(s,x)$ for $y^*$.
The optimization objective is
\begin{equation}
\begin{split}
  \mathcal{J}
  &= \max_{\pi_{\mathrm{ind}}}\;
    \mathbb{E}_{\tau,\,\mathcal{D}_\text{supp}^{(\tau)},\,(x,y^*)}
    \Bigl[ \\
  &\qquad
    \mathbb{E}_{s \sim \pi_{\mathrm{ind}}(\cdot \mid \mathcal{D}_\text{supp}^{(\tau)})}
    \bigl[\mathbf{1}\bigl[\hat{y}(s,x)=y^*\bigr]\bigr]
    \Bigr].
\end{split}
\label{eq:j_tir}
\end{equation}
\noindent where $\mathbf{1}[\cdot]$ is the indicator function, the outer expectation is over tasks, supports, and queries, and the inner expectation is over $s \sim \pi_{\mathrm{ind}}$.

% ============================================================
\section{CoDI}
\label{sec:method}

\noindent In this section, we present the proposed \codi{} for TIR-based inductive reasoning,
built on a cognitive-decoupling architecture (\S\ref{sec:arch}) and a dual-loop
meta-inductive learning scheme (\S\ref{sec:object}).

% \noindent We present \codi{}, a cognitively decoupled meta-inductive framework for TIR-based inductive reasoning. It consists of a cognitive-decoupling architecture (\S\ref{sec:arch})and a dual-loop meta-inductive learning objective (\S\ref{sec:object}).

\subsection{Overview}
\noindent The proposed \codi{} employs a cognitive-decoupling induction–execution framework: the query-blind inducer abstracts TIR and the executor solves new query based on it.
To enhance the inductive reasoning ability for general tasks, a dual-loop meta-inductive learning approach is proposed: the inner loop is executed based on TIR, and improves answers through ICL; the outer loop is based on the feedback signal from the inner loop execution, and optimizes the abstraction of TIR through RL.
This design is overviewed in Figure~\ref{fig:arch}.

% 为了提高通用领域任务的归纳推理能力，双层元归纳学习被提出：内环基于tir执行，通过ICL改善答案；外环基于内环执行的反馈信号，通过rl优化tir的抽取。

% To ensure that the TIR for specific task is reusable, 

% 为了保证特定任务生成可复用的TIR，cognitive-decoupling 归纳被提出，通过查询盲和结构化生成约束。

% \codi{} 将``先归纳、后执行''操作化为\textbf{query-blind}的梯度分区双层系统；图~\ref{fig:arch} 给出 (a)--(c) 总览。
% \textbf{\S\ref{sec:arch}（架构）}：\sg/\as{} 分工、两步接口与策略书写格式；\textbf{\S\ref{sec:object}（对象层）}：前向推理、梯度分区及以预测正确率为目标的 $\mathcal{J}_{\text{CoDI}}$；
% \textbf{\S\ref{sec:training}（元层训练）}：先 SFT 冷启动 \sg{}，再以 EIR/ICR 为信号做跨任务 GRPO。

% ============================================================
\subsection{Cognitive-Decoupling Framework}
\label{sec:arch}

\noindent \codi{} cognitively decouples induction and execution through two modules.
The Strategy Generator (\sg{}) induces TIR from few-shot demonstrations alone, and the Answer Solver (\as{}) answers each query by executing the TIR.

\paragraph{Inducer (\sg{}).}
The \sg{} induces a natural-language strategy $s$ from
$\mathcal{D}_\text{supp}^{(\tau)}$ alone, where $s$ serves as the TIR of
task $\tau$.
The query-blind interface isolates reusable task regularities from
query-specific cues.
Specifically, $s$ comprises \texttt{[INDUCTIVE\_RULE]},
\texttt{[TASK\_PROCEDURE]} (3--6 executable steps), and
\texttt{[TRANSFER\_SCOPE]} (Appendix~\ref{app:schema}).
Each section plays a distinct role:
\texttt{[INDUCTIVE\_RULE]} states verifiable task-family regularities;
\texttt{[TASK\_PROCEDURE]} operationalizes them as executable steps;
and \texttt{[TRANSFER\_SCOPE]} specifies applicability boundaries.
This semi-structured format yields comparable rollouts across candidates
and enables format validation during training.
The induction step is formalized as
\begin{equation}
  s = \mathrm{LLM}_{\mathrm{SG}}\!\bigl(\mathcal{D}_\text{supp}^{(\tau)}\bigr).
  \label{eq:sg_induce}
\end{equation}

\paragraph{Executor (\as{}).}
The \as{} applies $s$ to query $x$ via the concatenated prompt $(s,x)$.
It remains frozen throughout training; gradient partitioning blocks
back-propagation to \sg{}, so \sg{} is updated only through the discrete
strategy interface.
The execution step is formalized as
\begin{equation}
  \hat{y} = \mathrm{LLM}_{\mathrm{AS}}(s,\, x).
  \label{eq:as_exec}
\end{equation}
Gradient partitioning confines updates to $\pi_{\mathrm{SG}}$.
Without this barrier, gradients from query-conditioned execution would leak into the inducer and blur the TIR interface, conflating reusable task regularities with instance-specific matching.
Because \as{} remains frozen, execution accuracy reflects each TIR's transfer value rather than co-adaptation between modules.

% ============================================================
\subsection{Dual-Loop Meta-Inductive Learning}
\label{sec:object}

\noindent Building on the cognitive-decoupling framework, dual-loop meta-inductive learning evolves TIR when no strategy labels are available.
Execution under a frozen \as{} provides a weak, task-agnostic supervision signal for $\pi_{\mathrm{SG}}$.
$\pi_{\mathrm{SG}}$ is first warmed up via SFT on teacher demonstrations to satisfy the schema.
During RL, format-invalid rollouts receive zero reward and are excluded from \as{} execution.

\paragraph{Inner loop: TIR application.}
At each training step, \sg{} induces $G$ TIR candidates
$\{s_i\}_{i=1}^{G}$ from the support set $\mathcal{D}_{\text{supp}}^{(\tau)}$.
The frozen \as{} applies each $s_i$ to query $x$ through in-context prompting on $(s_i, x)$, without updating parameters.
This step tests whether each induced TIR generalizes beyond the few-shot demonstrations.
Execution yields a prediction and per-token log-probabilities;
the resultant accuracy $a_i$ and decoding distribution
serve as reward signals for the outer loop.

\paragraph{Outer loop: TIR evolution.}
The outer loop optimizes $\pi_{\mathrm{SG}}$ via cross-task GRPO
with two complementary rewards:

\begin{itemize}[leftmargin=*, itemsep=4pt]

\item \textbf{Execution Incremental Reward (EIR)}  measures
difficulty-adaptive accuracy gain.
Let $a_i\!\in\![0,1]$ be the answer accuracy of \as{} under $s_i$,
$\delta_i = a_i - b_\tau$ its deviation from a per-task cached
baseline $b_\tau$ (fixed before the $G$-rollout group; updated via
exponential moving average to prevent within-group leakage),
and $g_{\text{raw}}$ the difficulty-normalized gain
(log-smoothing constant $k$; Appendix~\ref{app:sensitivity}):
\begin{equation}
  \delta_i = a_i - b_\tau,\quad
  g_{\text{raw}} = \frac{\delta_i}{\max\!\bigl(
    b_\tau^{\alpha}(1{-}b_\tau)^{1-\alpha},\;\varepsilon\bigr)}
  \label{eq:eir_raw}
\end{equation}
where $\alpha\!=\!\alpha^{+}$ if $\delta_i\!\ge\!0$, else $\alpha^{-}$
($\alpha^{+}\!>\!\alpha^{-}$; Appendix~\ref{app:eir_normalizer}):
\begin{equation}
  R_{\text{EIR},i} = \operatorname{sign}(g_{\text{raw}})\cdot
    \ln\!\bigl(1+k\,|g_{\text{raw}}|\bigr).
  \label{eq:eir}
\end{equation}
For positive $\delta_i$, EIR sets $\alpha{=}\alpha^{+}$, down-weighting gains on easy tasks;
for negative $\delta_i$, it sets $\alpha{=}\alpha^{-}$, amplifying penalties on high-baseline regressions (Appendix~\ref{app:eir_normalizer}).

\item \textbf{Internal Confidence Reward (ICR)} measures the extent
to which $s_i$ concentrates \as{}'s decoding distribution,
distinguishing systematic accuracy from chance-level correctness.
$R_{\text{ICR},i}\!\in\!(0,1]$ is the geometric mean of per-token
probabilities under \as{}, normalized by output length $T_i$:
\begin{equation}
  R_{\text{ICR},i}
  = \exp\!\Bigl(\frac{1}{T_i}\sum_{t=1}^{T_i}
    \log \pas{}\bigl(y_t\mid y_{<t},s_i,x\bigr)\Bigr).
  \label{eq:icr}
\end{equation}

\end{itemize}

EIR and ICR are combined via multiplicative gating:
\begin{equation}
  R_i = R_{\text{EIR},i}\cdot\bigl(1+\beta\cdot R_{\text{ICR},i}\bigr),
  \label{eq:total_reward}
\end{equation}
where $\beta$ is set in Appendix~\ref{app:sensitivity}.
The factor $(1{+}\beta R_{\text{ICR},i})\!>\!0$ up-weights positive
EIR proportionally to prediction confidence without reversing
the sign of negative EIR.
EIR normalizes rewards by per-task difficulty so that $\pi_{\mathrm{SG}}$ is not biased toward easy tasks.
ICR complements EIR by up-weighting strategies whose frozen \as{} decoding is concentrated, separating systematic correctness from lucky hits.
Together, EIR and ICR supply complementary, label-free signals for meta-inductive training; the full SFT$\to$RL pipeline and hyperparameters are in Appendix~\ref{app:training}.

% ============================================================
\section{Experiments}
\label{sec:experiments}

\begin{table*}[!t]
  \centering
  \fontsize{6.8pt}{7.8pt}\selectfont
  \setlength{\tabcolsep}{3pt}
  \renewcommand{\arraystretch}{0.82}
  % ---------------------------------------------------------------------------
  % Column map (10 columns):
  %   c1 Type  c2 Model  c3 Method
  %   c4 BBH  c5 ID (BIG-Bench)  c6 OOD+Δ% (BIG-Bench)
  %   c7 MATH500  c8 StrategyQA  c9 ReClor  c10 Avg.\ (= mean of c4--c9)
  % 0S/S-0S rows omitted here for space.
  % MetaICL in w/ Training Qwen3-4B; GEPA on Qwen3-4B (Appendix~\ref{app:models}).
  % ---------------------------------------------------------------------------
  \resizebox{\textwidth}{!}{%
  \begin{tabular}{
    >{\centering\arraybackslash}p{0.75cm}%   c1 Type (rotated label)
    >{\raggedright\arraybackslash}p{1.5cm}%  c2 Model
    @{\hskip 12pt}%                          extra gap: Model → Method
    >{\raggedright\arraybackslash}p{1.2cm}%   c3 Method
    S[table-format=2.1]%                     c4 BBH
    S[table-format=2.1]%                     c5 BIG-Bench ID
    >{\centering\arraybackslash}p{1.35cm}%   c6 BIG-Bench OOD+Δ%
    >{\centering\arraybackslash}p{1.1cm}%     c7 MATH500
    >{\centering\arraybackslash}p{1.1cm}%    c8 StrategyQA
    >{\centering\arraybackslash}p{1.0cm}%    c9 ReClor
    >{\centering\arraybackslash}p{1.0cm}%    c10 Avg.
  }
    \toprule
    & & & & \multicolumn{2}{c}{\textbf{BIG-Bench}} & & & & \\[-2pt]
    \cmidrule(lr){5-6}
    \multicolumn{1}{c}{\textbf{Type}} &
    \multicolumn{1}{c}{\textbf{Model}} &
    \multicolumn{1}{c}{\textbf{Method}} &
    \multicolumn{1}{c}{\textbf{BBH} ($\Delta$\%)} &
    \multicolumn{1}{c}{\textbf{ID} ($\Delta$\%)} &
    \multicolumn{1}{c}{\textbf{OOD} ($\Delta$\%)} &
    \multicolumn{1}{c}{\textbf{MATH-500} ($\Delta$\%)} &
    \multicolumn{1}{c}{\textbf{StratQA} ($\Delta$\%)} &
    \multicolumn{1}{c}{\textbf{ReClor} ($\Delta$\%)} &
    \multicolumn{1}{c}{\textbf{Avg.}} \\
    \midrule
    %
    % ══════════════════════════════════════════════════════════
    % BLOCK 1: w/o Training  (12 rows)
    %   Gemma-3-4B(2) + Qwen3-4B(3) + Llama-3.1-8B(2)
    %   + Gemini-3-flash†(3) + GPT-5†(2)
    %   [Qwen3-8B omitted from main table for space]
    % ══════════════════════════════════════════════════════════
    \multirow{12}{*}{\rotatebox{90}{\textit{w/o Training}}}
    % ── Gemma-3-4B ────────────────────────────── 2 rows ──
      & \multirow{2}{*}{Gemma-3-4B}
        & FS   & 37.41 & 26.00 & 45.00 & 54.00 & 58.95 & 48.00 & 44.89 \\
    & & S-FS & \multicolumn{1}{c}{25.00\,\dg{-33}} & \multicolumn{1}{c}{18.33\,\dg{-30}} & 28.00\,\dg{-38} & 60.00\,\dr{11.1} & 59.97\,\dr{1.7} & 46.40\,\dg{-3.3} & 39.62 \\
    \cmidrule(l){2-10}
    % ── Qwen3-4B: FS, S-FS, GEPA ─────────────── 3 rows ──
      & \multirow{3}{*}{Qwen3-4B}
        & FS   & 35.56 & 29.33 & 53.33 & 63.13 & 69.97 & 83.60 & 55.82 \\
    & & S-FS & \multicolumn{1}{c}{27.78\,\dg{-22}} & \multicolumn{1}{c}{30.00\,\dr{2}} & 38.33\,\dg{-28} & 65.20\,\dr{3.3} & 64.19\,\dg{-8.3} & 81.20\,\dg{-2.9} & 51.12 \\
    & & GEPA$^{\S}$ & {79.59} & {61.65} & {56.88} & {61.20} & {71.32} & {79.80} & {68.41} \\
    \cmidrule(l){2-10}
    % ── Llama-3.1-8B ──────────────────────────── 2 rows ──
      & \multirow{2}{*}{Llama-3.1-8B}
        & FS   & 31.11 & 20.00 & 44.33 & 43.40 & 70.60 & 39.40 & 41.47 \\
    & & S-FS & \multicolumn{1}{c}{21.30\,\dg{-32}} & \multicolumn{1}{c}{13.33\,\dg{-33}} & 23.00\,\dg{-48} & 37.80\,\dg{-12.9} & 53.13\,\dg{-24.7} & 40.60\,\dr{3.0} & 31.53 \\
    \cmidrule(l){2-10}
    % ── Gemini-3-flash†: FS, S-FS ────────────── 2 rows ──
      & \multirow{2}{*}{Gemini-3-flash$^{\dagger}$}
        & FS   & 95.37 & 84.67 & 84.00 & 65.93 & 83.50 & 84.53 & 83.00 \\
    & & S-FS & \multicolumn{1}{c}{91.67\,\dg{-4}} & \multicolumn{1}{c}{76.00\,\dg{-10}} & 78.67\,\dg{-6} & 60.60\,\dg{-8.1} & 87.29\,\dr{4.5} & 85.00\,\dr{0.6} & 79.87 \\
    \cmidrule(l){2-10}
    % ── GPT-5†: FS, S-FS ─────────────────────────── 2 rows ──
      & \multirow{2}{*}{GPT-5$^{\dagger}$}
        & FS   & 94.44 & 80.33 & 73.67 & 71.40 & 85.54 & 97.93 & 83.89 \\
    & & S-FS & \multicolumn{1}{c}{88.89\,\dg{-6}} & \multicolumn{1}{c}{63.00\,\dg{-22}} & 61.67\,\dg{-16} & 68.00\,\dg{-4.8} & 97.93\,\dr{14.5} & 98.00\,\dr{0.1} & 79.58 \\
    %
    % ══════════════════════════════════════════════════════════
    % BLOCK 2: w/ Training  (7 rows)
    %   Gemma-3-4B(2) + Qwen3-4B(3) + Llama-3.1-8B(2)
    %   [Qwen3-8B omitted from main table for space]
    % Cols 1–2 (Type, Model) uncolored so \multirow text is unobstructed.
    % Δ% on MIST OOD = (w/ MIST OOD − w/o FS OOD) / w/o FS OOD × 100
    % ══════════════════════════════════════════════════════════
    \midrule
    \multirow{7}{*}{\rotatebox{90}{\textit{w/ Training}}}
    % ── Gemma-3-4B ────────────────────────────── 2 rows ──
      & \multirow{2}{*}{Gemma-3-4B}
        & \cellcolor{rlbg}FS   & \multicolumn{1}{c}{\cellcolor{rlbg}40.00\,\dr{7}} & \multicolumn{1}{c}{\cellcolor{rlbg}31.33\,\dr{21}} & \cellcolor{rlbg}50.33\,\dr{12} & \cellcolor{rlbg}61.60\,\dr{14.1} & \cellcolor{rlbg}\textbf{69.72}\,\dr{18.3} & \cellcolor{rlbg}49.20\,\dr{2.5} & \cellcolor{rlbg}50.36 \\
    & & \cellcolor{mist}\gist{} & \multicolumn{1}{c}{\cellcolor{mist}\textbf{42.67}\,\dr{14}} & \multicolumn{1}{c}{\cellcolor{mist}\textbf{43.00}\,\dr{65}} & \cellcolor{mist}\textbf{55.67}\,\dr{24} & \cellcolor{mist}\textbf{62.20}\,\dr{15.2} & \cellcolor{mist}67.98\,\dr{15.3} & \cellcolor{mist}\textbf{65.60}\,\dr{36.7} & \cellcolor{mist}\textbf{56.19} \\
    \cmidrule(l){2-10}
    % ── Qwen3-4B: FS, GIST, MetaICL ─────────────── 3 rows ──
      & \multirow{3}{*}{Qwen3-4B}
        & \cellcolor{rlbg}FS   & \multicolumn{1}{c}{\cellcolor{rlbg}40.93\,\dr{15}} & \multicolumn{1}{c}{\cellcolor{rlbg}41.33\,\dr{41}} & \cellcolor{rlbg}55.56\,\dr{4.2} & \cellcolor{rlbg}65.80\,\dr{4.2} & \cellcolor{rlbg}73.53\,\dr{5.1} & \cellcolor{rlbg}85.80\,\dr{2.6} & \cellcolor{rlbg}60.49 \\
    & & \cellcolor{mist}\gist{} & \multicolumn{1}{c}{\cellcolor{mist}\textbf{82.22}\,\dr{131}} & \multicolumn{1}{c}{\cellcolor{mist}\textbf{64.67}\,\dr{120}} & \cellcolor{mist}\textbf{63.00}\,\dr{18} & \cellcolor{mist}\textbf{76.15}\,\dr{20.6} & \cellcolor{mist}\textbf{77.44}\,\dr{10.7} & \cellcolor{mist}\textbf{86.60}\,\dr{3.6} & \cellcolor{mist}\textbf{75.01} \\
    & & MetaICL & 33.52 & 32.00 & 28.67 & 16.80 & 58.95 & 72.00 & 40.32 \\
    \cmidrule(l){2-10}
    % ── Llama-3.1-8B ──────────────────────────── 2 rows ──
      & \multirow{2}{*}{Llama-3.1-8B}
        & \cellcolor{rlbg}FS   & \multicolumn{1}{c}{\cellcolor{rlbg}32.22\,\dr{4}} & \multicolumn{1}{c}{\cellcolor{rlbg}22.11\,\dr{11}} & \cellcolor{rlbg}45.56\,\dr{3} & \cellcolor{rlbg}\textbf{50.00}\,\dr{15.2} & \cellcolor{rlbg}\textbf{81.51}\,\dr{15.5} & \cellcolor{rlbg}45.60\,\dr{15.7} & \cellcolor{rlbg}46.17 \\
    & & \cellcolor{mist}\gist{} & \multicolumn{1}{c}{\cellcolor{mist}\textbf{41.67}\,\dr{34}} & \multicolumn{1}{c}{\cellcolor{mist}\textbf{23.00}\,\dr{15}} & \cellcolor{mist}\textbf{48.33}\,\dr{9} & \cellcolor{mist}47.80\,\dr{10.1} & \cellcolor{mist}73.94\,\dr{4.7} & \cellcolor{mist}\textbf{48.00}\,\dr{21.8} & \cellcolor{mist}\textbf{47.12} \\
    \bottomrule
  \end{tabular}%
  }% end \resizebox
  \caption{Exact-match accuracy comparing \gist{} with FS, S-FS, GEPA, MetaICL, and closed-source APIs on BBH, BIG-Bench ID/OOD, MATH-500, StrategyQA, and ReClor (w/o and w/ training).
  $\Delta$\%$=(\text{score}-\text{FS})/\text{FS}\times100$; \textcolor{BrickRed}{red}/\textcolor{NavyBlue}{blue} = gain/loss.
  $^{\dagger}$Unfine-tuned; Gemini-3-flash = \texttt{gemini-3-flash-preview}.
  $^{\S}$GEPA: \texttt{gemini-3-flash-preview} optimizer, Qwen3-4B executor, four reflective iterations.}
  \label{tab:main_results}
\end{table*}

\subsection{Experimental Setup}
\label{sec:setup}

\paragraph{Baselines.}
\textbf{FS} (few-shot ICL) prepends $K$ demonstrations for decoding~\citep{dong2024icl}.
\textbf{S-FS} (strategy-based few-shot; untrained-inducer baseline) induces a strategy before straight answering~\citep{gao2024neurips-strategyllm}, sharing the query-blind interface without cross-task RL.
\textbf{MetaICL}~\citep{min2022metaicl} performs cross-task meta-training under equal compute without explicit induction.
\textbf{GEPA}~\citep{agrawal2026gepareflectivepromptevolution} performs inference-time prompt optimization.
\textbf{Gemini-3-flash} and \textbf{GPT-5} are closed-source upper bounds.
See Appendix~\ref{app:training} for details.

\paragraph{Datasets and Metrics.}
Training data are drawn from \textbf{StrategyBench}~\citep{anonymous_strategybench}, a curated subset of \textbf{BIG-Bench}~\citep{srivastava2023beyond} filtered according to five inducibility criteria (Appendix~\ref{app:filtering}) using Qwen3-32B scoring with author spot-checks.
We use execution-filtered demonstrations from this corpus for SFT (Appendix~\ref{app:strategybench}).
Evaluation covers \textbf{BIG-Bench} ID/OOD splits (Appendix~\ref{app:ood_partition}), \textbf{BBH}~\citep{suzgun2022bbh}, \textbf{MATH-500}~\citep{lightman2023lets}, \textbf{StrategyQA}~\citep{geva2021strategyqa}, and \textbf{ReClor}~\citep{yu2020reclor}, with exact-match accuracy throughout.
$\Delta$\% denotes relative change from the same-backbone FS baseline: $\Delta = (\text{score} - \text{FS}) / \text{FS} \times 100\%$; absolute accuracy differences are reported in percentage points (pp); Avg.\ in Table~\ref{tab:main_results} averages the five benchmarks.

% ----------------------------------------------------------
\subsection{Main Results}
\label{sec:main_results}

% Table~\ref{tab:main_results} compares \gist{} with ICL and meta-induction baselines on six benchmarks.

\noindent To isolate what enables transferable TIR, we compare methods in Table~\ref{tab:main_results} and obtain three findings.

\paragraph{Cross-task training alone cannot yield TIR.}
Under equal training budgets, MetaICL scores below the untrained FS baseline on OOD and trails \gist{} by over 34 points, showing cross-task parameter updates alone fail to produce reusable task-level strategies.
GEPA improves over FS on in-distribution tasks, but its gains derive from query-conditioned prompt evolution rather than query-decoupled strategy induction, placing it structurally apart from the reusable TIR that \gist{} targets.

\paragraph{Query-blind interface alone is insufficient.}
To disentangle interface from meta-inductive training, we compare S-FS with FS under the same backbone.
S-FS retains \gist{}'s query-blind interface but excludes cross-task RL.
On OOD, S-FS trails FS across all three backbones and remains far below trained \gist{}, showing query-blind decoupling alone is insufficient for generalizable TIR.

\paragraph{Explicit TIR-aligned training drives gains.}
\gist{} consistently outperforms all baselines under full TIR-aligned training.
On Qwen3-4B, it raises OOD accuracy by $+18\%$ over the untrained FS baseline, with especially pronounced gains on BBH and BIG-Bench ID.
Positive OOD improvements hold across all three open-source backbones and extend to MATH-500, StrategyQA, and ReClor.
These results confirm that optimizing explicitly for TIR generalizes task-level strategy induction across diverse backbones and domains.
% Relative ID gains exceed OOD gains, as expected under ID-only RL supervision with held-out subtasks at test time; the consistent OOD improvements across backbones nonetheless indicate generalization beyond memorized training tasks.
% ----------------------------------------------------------
\subsection{Ablation Study}
\label{sec:ablation}

\noindent To verify what drives generalizable TIR, we ablate core components on Qwen3-4B (Table~\ref{tab:ablation},~\ref{tab:ablation_format}).

\paragraph{Reward design and training pipeline.}
We analyze the effect of reward design, training stages, and architectural decoupling on Qwen3-4B.
As shown in Table~\ref{tab:ablation}, the full model achieves
63.00 OOD accuracy.
Removing ICR preserves ID but reduces OOD by 9.3~pp, indicating ICR supports cross-task generalization.
Without EIR, ID falls to 49.00, as rewards without difficulty-adaptive normalization conflate task difficulty with strategy quality.
Raw-accuracy and LLM-judge rewards, SFT-only training, and end-to-end
non-decoupled training all trail substantially, indicating that neither alternative rewards nor
query-coupled optimization suffices for generalizable TIR induction.

\begin{table}[h!]
  \centering
  \fontsize{7.5pt}{9.5pt}\selectfont
  \setlength{\tabcolsep}{4pt}
  \renewcommand{\arraystretch}{0.95}
  \resizebox{\columnwidth}{!}{%
  \begin{tabular}{@{}l S[table-format=2.2] S[table-format=2.2] S[table-format=2.2]@{}}
    \toprule
    \textbf{Configuration} &
    {\textbf{ID}} & {\textbf{BBH}} & {\textbf{OOD}} \\
    \midrule
    \rowcolor{mist}
    \gist{} (EIR+ICR, full)  & \textbf{64.67} & \textbf{82.22} & \textbf{63.00} \\
    \midrule
    \multicolumn{4}{@{}l}{\textit{Reward component ablation}} \\
    \quad w/o ICR ($\beta{=}0$, EIR only)  & 62.67 & 69.81 & 53.67 \\
    \quad w/o EIR (ICR only)               & 49.00 & 69.07 & 53.67 \\
    \quad w/ raw Acc reward                & 31.67 & 48.15 & 56.67 \\
    \quad w/ LLM Judge reward              & 49.00 & 69.44 & 57.33 \\
    \midrule
    \multicolumn{4}{@{}l}{\textit{Training pipeline ablation}} \\
    \quad w/o SFT (RL only)               & 60.67 & 78.70 & 55.00 \\
    \quad w/o RL (SFT only)               & 31.67 & 33.89 & 40.00 \\
    \midrule
    \multicolumn{4}{@{}l}{\textit{Decoupled architecture}} \\
    \quad w/o decoupled arch.\ (e2e training)  & 33.67 & 38.52 & 40.33 \\
    \bottomrule
  \end{tabular}%
  }
  \caption{Qwen3-4B component ablations.}
  \label{tab:ablation}
\end{table}

\paragraph{Strategy format.}
Table~\ref{tab:ablation_format} compares TIR formats; the semi-structured schema outperforms free-form and CoT-style alternatives on ID, BBH, OOD, and ICR.
Removing \texttt{[TASK\_PROCEDURE]} collapses OOD to 2.67, the largest impact among its three sections; ablating \texttt{[INDUCTIVE\_RULE]} or \texttt{[TRANSFER\_SCOPE]} lowers OOD by over 20~pp each, showing that both the induced rule and transfer scope are required for generalization.

\begin{table}[h!]
  \centering
  \fontsize{7.5pt}{9.5pt}\selectfont
  \setlength{\tabcolsep}{4pt}
  \renewcommand{\arraystretch}{0.95}
  \resizebox{\columnwidth}{!}{%
  \begin{tabular}{@{}l S[table-format=2.2] S[table-format=2.2] S[table-format=2.2] S[table-format=2.2]@{}}
    \toprule
    \textbf{Strategy Format} &
    {\textbf{ID}} & {\textbf{BBH}} & {\textbf{OOD}} &
    {\textbf{ICR}} \\
    \midrule
    \rowcolor{mist}
    \gist{} (semi-structured)  & \textbf{64.67} & \textbf{82.22} & \textbf{63.00} & \textbf{84.73} \\
    \midrule
    Free-form (no schema)               & 46.00 & 53.52 & 40.00 & 75.59 \\
    CoT-style (implicit structure)      & 59.56 & 72.84 & 49.33 & 75.57 \\
    \midrule
    \multicolumn{5}{@{}l}{\textit{Semi-structured section ablation}} \\
    \quad w/o \texttt{[INDUCTIVE\_RULE]}   & 42.33 & 49.44 & 39.67 & 74.27 \\
    \quad w/o \texttt{[TASK\_PROCEDURE]}   & 1.33 & 1.48 & 2.67 & 71.67 \\
    \quad w/o \texttt{[TRANSFER\_SCOPE]}   & 31.67 & 36.85 & 33.67 & 80.35 \\
    \bottomrule
  \end{tabular}%
  }
  \caption{Qwen3-4B ablation comparing alternative TIR formats and individual schema sections.}
  \label{tab:ablation_format}
\end{table}

% ============================================================
\section{Analysis}
\label{sec:analysis}

% ============================================================
\subsection{What Drives the Gains?}
\label{sec:attribution}

% \noindent These analyses address Intro claims that TIR quality is independently optimizable from scale and that meta-induction requires cross-example abstraction rather than instance memorization.

\begin{figure}[h]
  \centering
  \begin{tikzpicture}
    \begin{axis}[
      width  = 0.82\columnwidth,
      height = 4.2cm,
      xlabel = {SG Parameter Count (B)},
      ylabel = {OOD Accuracy},
      xlabel style    = {font=\small, color=black!75, yshift=-1pt},
      ylabel style    = {font=\small, color=black!75, xshift=-2pt},
      xticklabel style = {font=\footnotesize, color=black!65},
      yticklabel style = {font=\footnotesize, color=black!65},
      xmin = 1.0, xmax = 36,
      ymin = 0.10, ymax = 0.67,
      xtick = {1.7, 4, 8, 14, 32},
      xticklabels = {1.7, 4, 8, 14, 32},
      ytick = {0.1, 0.2, 0.3, 0.4, 0.5, 0.6},
      grid = both,
      grid style      = {gray!18, densely dashed, thin},
      minor tick num  = 0,
      axis line style = {black!35, line width=0.5pt},
      tick style      = {black!35, line width=0.5pt, major tick length=3pt},
      clip = false,
      %% legend: top-center, horizontal, no border/background
      legend columns   = 2,
      legend cell align = left,
      legend style = {
        at           = {(0.50, 1.04)},
        anchor       = south,
        font         = \small,
        draw         = none,
        fill         = none,
        column sep   = 16pt,
        row sep      = 0pt,
        inner sep    = 0pt,
      },
    ]
      %% shaded area under scaling curve
      \addplot[draw=none, fill=NavyBlue!9, forget plot]
        coordinates {(1.7,0.10)(1.7,0.1200)(4,0.3833)(8,0.4500)(14,0.5300)(32,0.5684)(32,0.10)}
        \closedcycle;
      %% Untuned SG scaling curve  (dashed + filled dots, matching reference style)
      \addplot[
        NavyBlue!85!black, line width=1.5pt, densely dashed,
        mark=*, mark size=3.4pt,
        mark options={fill=NavyBlue!85!black, draw=white, line width=0.8pt},
      ] coordinates {(1.7,0.1200)(4,0.3833)(8,0.4500)(14,0.5300)(32,0.5684)};
      \addlegendentry{Untuned \sg{}}
      %% GIST-4B reference dashed line (full width)
      \addplot[
        MediumPurple!55, line width=1.0pt, densely dashed, forget plot,
      ] coordinates {(1.0, 0.6193)(36, 0.6193)};
      \addlegendimage{
        MediumPurple!90!black, only marks, mark=asterisk,
        mark size=7.0pt, line width=1.1pt,
      }
      \addlegendentry{\gist{}-4B}
      %% GIST-4B star marker
      \addplot[
        white, only marks, mark=asterisk, mark size=7.6pt,
        line width=2.6pt, forget plot,
      ] coordinates {(4, 0.6193)};
      \addplot[
        MediumPurple!90!black, only marks, mark=asterisk, mark size=7.0pt,
        line width=1.1pt, forget plot,
      ] coordinates {(4, 0.6193)};
    \end{axis}
  \end{tikzpicture}
  \caption{Trained \gist{}-4B exceeds untuned 32B \sg{} on BIG-Bench OOD with $8\times$ fewer parameters.}
  \label{fig:scale_sweep_main}
\end{figure}
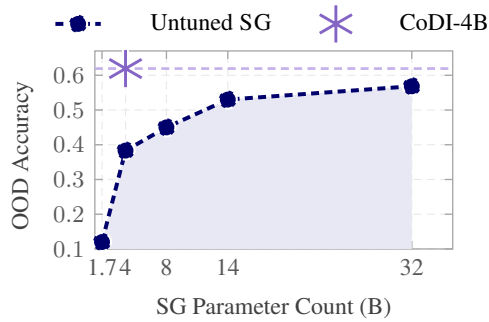

\paragraph{Gains exceed scale.}
Analysis reveals that OOD improvements primarily reflect TIR optimization rather than pretrained capacity alone.
With \as{} fixed to Qwen3-4B, untuned \sg{} OOD accuracy rises from 0.12 (1.7B) to 0.57 (32B), confirming that scale remains helpful.
Trained \gist{}-4B nonetheless reaches 0.62 and exceeds the 32B baseline by 9\% despite $8\times$ fewer parameters (Figure~\ref{fig:scale_sweep_main}).
Because the gap persists after untuned scaling saturates, TIR quality functions as a distinct optimization target orthogonal to model size.

% Figure 3: scale sweep

\begin{figure}[h]
  \centering
  \definecolor{figfourpurple}{HTML}{3B168B}
  \definecolor{figfourlight}{HTML}{F4F0FB}
  \definecolor{figfourgrid}{HTML}{FFFFFF}
  \definecolor{figfourmuted}{HTML}{6E5F81}
  \definecolor{figfourink}{HTML}{1E1733}
  \resizebox{0.99\columnwidth}{!}{%
  \begin{tikzpicture}[font=\scriptsize, x=1cm, y=1cm]
    \pgfmathsetmacro{\cw}{0.82}
    \pgfmathsetmacro{\ch}{0.36}
    \pgfmathsetmacro{\labelX}{0.43}
    \pgfmathsetmacro{\preX}{0.62}
    \pgfmathsetmacro{\postX}{3.88}
    \pgfmathsetmacro{\cbX}{6.94}
    \pgfmathsetmacro{\topY}{0.00}
    \pgfmathsetmacro{\midX}{(\preX+3*\cw+\postX)/2}

    \node[anchor=south, align=center, font=\scriptsize\bfseries, text=figfourpurple]
      at ({\preX+1.5*\cw},0.70) {Pre-\gist{}\\{\normalfont\fontsize{5.3pt}{5.7pt}\selectfont(base model)}};
    \node[anchor=south, align=center, font=\scriptsize\bfseries, text=figfourpurple]
      at ({\postX+1.5*\cw},0.70) {Post-\gist{}\\{\normalfont\fontsize{5.3pt}{5.7pt}\selectfont(\gist{} fine-tuned)}};
    \draw[-{Stealth[length=4pt,width=5pt]}, figfourpurple, line width=0.45pt]
      ({\preX+3*\cw-0.38},1.22) -- ({\postX+0.40},1.22);
    \node[font=\footnotesize\bfseries, text=figfourink] at (\midX,1.39) {$-73.4\%$};

    \foreach \i/\lab in {0/Ex1,1/Ex2,2/Ex3} {
      \node[font=\scriptsize\bfseries, anchor=south]
        at ({\preX+(\i+0.5)*\cw},0.32) {\lab};
      \node[font=\scriptsize\bfseries, anchor=south]
        at ({\postX+(\i+0.5)*\cw},0.32) {\lab};
    }

    \foreach \r/\task/\a/\b/\c/\d/\e/\f in {
      0/T01/0.459/0.445/0.500/0.082/0.088/0.085,
      1/T02/0.452/0.500/0.455/0.170/0.176/0.181,
      2/T03/0.353/0.377/0.364/0.081/0.080/0.088,
      3/T04/0.288/0.295/0.304/0.054/0.051/0.060,
      4/T05/0.281/0.275/0.302/0.065/0.070/0.062,
      5/T06/0.325/0.304/0.357/0.117/0.122/0.114,
      6/T07/0.247/0.265/0.232/0.039/0.039/0.039,
      7/T08/0.282/0.290/0.272/0.087/0.083/0.089,
      8/T09/0.248/0.245/0.245/0.071/0.064/0.073,
      9/T10/0.270/0.255/0.273/0.093/0.097/0.082} {
      \pgfmathsetmacro{\rowY}{\topY-\r*\ch}
      \node[anchor=east, font=\fontsize{5.2pt}{5.7pt}\selectfont\bfseries, text=figfourmuted]
        at (\labelX,{\rowY-0.5*\ch}) {\task};
      \foreach \i/\val in {0/\a,1/\b,2/\c} {
        \pgfmathsetmacro{\clipval}{min(\val,0.5)}
        \pgfmathtruncatemacro{\pct}{round(7+\clipval/0.5*66)}
        \fill[figfourpurple!\pct!white]
          ({\preX+\i*\cw},\rowY) rectangle ++(\cw,-\ch);
        \draw[figfourgrid, line width=0.45pt]
          ({\preX+\i*\cw},\rowY) rectangle ++(\cw,-\ch);
        \node[font=\fontsize{5.3pt}{5.8pt}\selectfont\bfseries, text=figfourink]
          at ({\preX+(\i+0.5)*\cw},{\rowY-0.5*\ch}) {\val};
      }
      \foreach \i/\val in {0/\d,1/\e,2/\f} {
        \pgfmathsetmacro{\clipval}{min(\val,0.5)}
        \pgfmathtruncatemacro{\pct}{round(7+\clipval/0.5*66)}
        \fill[figfourpurple!\pct!white]
          ({\postX+\i*\cw},\rowY) rectangle ++(\cw,-\ch);
        \draw[figfourgrid, line width=0.45pt]
          ({\postX+\i*\cw},\rowY) rectangle ++(\cw,-\ch);
        \node[font=\fontsize{5.3pt}{5.8pt}\selectfont\bfseries, text=figfourink]
          at ({\postX+(\i+0.5)*\cw},{\rowY-0.5*\ch}) {\val};
      }
    }

    \node[font=\scriptsize, align=center, anchor=south] at ({\cbX+0.22},0.31) {LOO\\sensitivity};
    \foreach \k/\v in {0/0.50,1/0.45,2/0.40,3/0.35,4/0.30,5/0.25,6/0.20,7/0.15,8/0.10,9/0.05} {
      \pgfmathtruncatemacro{\pct}{round(7+\v/0.5*66)}
      \fill[figfourpurple!\pct!white] (\cbX,{-\k*\ch-0.25}) rectangle ++(0.22,-\ch);
    }
    \foreach \tick/\lab in {0/0.5,2/0.4,4/0.3,6/0.2,8/0.1,10/0.0} {
      \draw[figfourmuted!45, line width=0.3pt]
        ({\cbX+0.22},{-\tick*\ch-0.25}) -- ({\cbX+0.28},{-\tick*\ch-0.25});
      \node[anchor=west, font=\scriptsize, text=figfourink]
        at ({\cbX+0.32},{-\tick*\ch-0.25}) {\lab};
    }

    \pgfmathsetmacro{\meanY}{-10*\ch-0.12}
    \foreach \x/\txt in {\preX/{mean = 0.325},\postX/{mean = 0.087}} {
      \fill[figfourlight] (\x,\meanY) rectangle ++({3*\cw},-0.32);
      \draw[figfourpurple, line width=0.35pt, rounded corners=1pt]
        (\x,\meanY) rectangle ++({3*\cw},-0.32);
      \node[font=\scriptsize\bfseries, text=figfourink]
        at ({\x+1.5*\cw},{\meanY-0.16}) {\txt};
    }
  \end{tikzpicture}%
  }
  \caption{Mean LOO rule sensitivity drops 73.4\% after training, supporting cross-example abstraction.}
  \label{fig:induction_mechanism}
\end{figure}
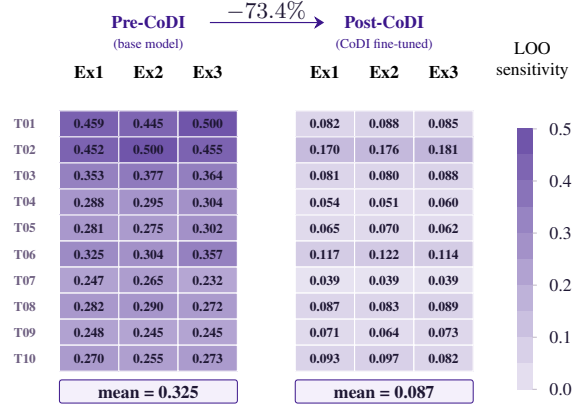

\paragraph{Induction becomes cross-example.}
We next test whether \gist{} induces rules from multiple supports rather than copying individual examples.
If rules still depended on single demonstrations, leave-one-out (LOO) removal should strongly alter \texttt{[INDUCTIVE\_RULE]} semantics.
On matched evaluation tasks, we regenerate each induced rule after removing one of $K{=}3$ supports and measure embedding shift (Appendix~\ref{app:loo_rule_diagnostics}).
Mean LOO sensitivity drops from 0.325 to 0.087 across 10 task types (Figure~\ref{fig:induction_mechanism}).
Accordingly, the gains reflect cross-example abstraction rather than instance memorization.

% Figure 4: LOO sensitivity heatmap (10 tasks, real run)

% ============================================================
\subsection{What Does Meta-Induction Reshape?}
\label{sec:reasoning_process}

% \noindent These analyses test whether meta-inductive training shifts models from query-coupled surface matching toward the decoupled, rule-directed induction promised in \S\ref{sec:intro}.

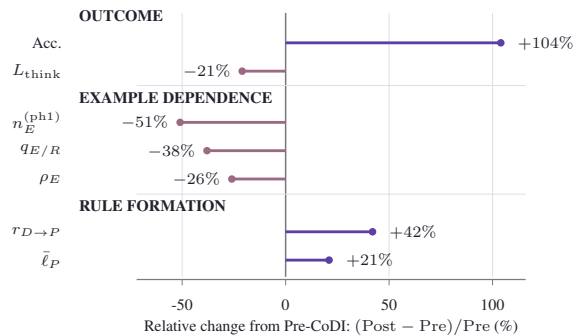
\begin{figure}[H]
  \centering
  \resizebox{0.994\columnwidth}{!}{%
  \begin{tikzpicture}[x=0.032cm, y=-0.60cm]
    \definecolor{fig5gain}{HTML}{5B3FA3}
    \definecolor{fig5drop}{HTML}{9A6A82}
    \definecolor{fig5ink}{HTML}{251F2B}
    \definecolor{fig5grid}{HTML}{E1DDE8}
    \def\xmin{-72}
    \def\xmax{116}
    \def\labX{-104}

    % Light guides and a stronger zero reference line.
    \foreach \xv in {-50,50,100} {
      \draw[fig5grid, line width=0.35pt] (\xv,-0.62) -- (\xv,6.18);
    }
    \draw[fig5ink!62, line width=0.65pt] (0,-0.62) -- (0,6.18);
    \draw[fig5grid, line width=0.4pt] (\xmin,1.25) -- (\xmax,1.25);
    \draw[fig5grid, line width=0.4pt] (\xmin,4.05) -- (\xmax,4.05);

    % Group headings and compact mathematical metric labels.
    \node[anchor=west, font=\scriptsize\bfseries, text=fig5ink]
      at (\labX,-0.55) {OUTCOME};
    \node[anchor=east, font=\scriptsize, text=fig5ink] at (\labX,0.15) {Acc.};
    \node[anchor=east, font=\scriptsize, text=fig5ink] at (\labX,0.88) {$L_{\mathrm{think}}$};

    \node[anchor=west, font=\scriptsize\bfseries, text=fig5ink]
      at (\labX,1.55) {EXAMPLE DEPENDENCE};
    \node[anchor=east, font=\scriptsize, text=fig5ink] at (\labX,2.20) {$n_E^{(\mathrm{ph1})}$};
    \node[anchor=east, font=\scriptsize, text=fig5ink] at (\labX,2.93) {$q_{E/R}$};
    \node[anchor=east, font=\scriptsize, text=fig5ink] at (\labX,3.66) {$\rho_E$};

    \node[anchor=west, font=\scriptsize\bfseries, text=fig5ink]
      at (\labX,4.35) {RULE FORMATION};
    \node[anchor=east, font=\scriptsize, text=fig5ink] at (\labX,5.02) {$r_{D\to P}$};
    \node[anchor=east, font=\scriptsize, text=fig5ink] at (\labX,5.75) {$\bar{\ell}_P$};

    % Signed lollipops: hue and sign both encode direction.
    \foreach \y/\v/\txt in {0.15/104/{+104\%},5.02/42/{+42\%},5.75/21/{+21\%}} {
      \draw[fig5gain, line width=1.25pt] (0,\y) -- (\v,\y);
      \fill[fig5gain] (\v,\y) circle[radius=1.65pt];
      \node[anchor=west, font=\scriptsize, text=fig5ink] at ({\v+4},\y) {$\txt$};
    }
    \foreach \y/\v/\txt in {0.88/-21/{-21\%},2.20/-51/{-51\%},2.93/-38/{-38\%},3.66/-26/{-26\%}} {
      \draw[fig5drop, line width=1.25pt] (0,\y) -- (\v,\y);
      \fill[fig5drop] (\v,\y) circle[radius=1.65pt];
      \node[anchor=east, font=\scriptsize, text=fig5ink] at ({\v-1.5},\y) {$\txt$};
    }

    % Bottom axis.
    \draw[fig5ink!60, line width=0.55pt] (\xmin,6.25) -- (\xmax,6.25);
    \foreach \xv in {-50,0,50,100} {
      \draw[fig5ink!60, line width=0.4pt] (\xv,6.25) -- (\xv,6.40);
      \node[anchor=north, font=\scriptsize, text=fig5ink] at (\xv,6.53) {\xv};
    }
    \node[anchor=north, font=\scriptsize, text=fig5ink] at ({(\xmin+\xmax)/2},7.05)
      {Relative change from Pre-\gist{}: $(\mathrm{Post}-\mathrm{Pre})/\mathrm{Pre}$ (\%)};
  \end{tikzpicture}%
  }
  \caption{Post-\gist{} raises OOD accuracy while shortening traces and example recitation, refuting a length-driven account and supporting rule-directed reasoning.}
  \label{fig:cognitive_trace_shift}
\end{figure}

\begin{figure}[H]
  \centering
  \begin{tikzpicture}
  \begin{axis}[
    xlabel          = {Number of shots $K$},
    ylabel          = {OOD Accuracy (\%)},
    xmin            = 0.55,
    xmax            = 5.45,
    ymin            = 38.5,
    ymax            = 57.5,
    xtick           = {1, 2, 3, 5},
    ytick           = {40, 44, 48, 52, 56},
    yticklabel style = {font=\scriptsize},
    x tick label style = {font=\scriptsize},
    ylabel style   = {font=\small},
    xlabel style   = {font=\small},
    width          = 0.90\columnwidth,
    height         = 5.05cm,
    axis line style = {gray!65},
    tick style     = {gray!65},
    legend style   = {at={(0.5,1.12)}, anchor=south,
                      font=\scriptsize,
                      legend columns=2,
                      legend cell align=left,
                      inner xsep=2pt,
                      inner ysep=1pt,
                      draw=none, fill=none,
                      /tikz/every even column/.append style={column sep=7pt}},
    grid           = major,
    grid style     = {gray!22, dashed},
    clip           = false,
  ]
  \addplot[red!72, thick, mark=square*, mark size=2.7pt,
           mark options={fill=red!75}]
    coordinates {(1,40.67) (2,53.00) (3,53.33) (5,53.00)};
  \addlegendentry{Few-shot ICL}
  \addplot[MediumPurple, very thick, mark=star, mark size=4.0pt,
           mark options={fill=MediumPurple}]
    coordinates {(1,52.67) (2,54.00) (3,55.56) (5,54.33)};
  \addlegendentry{\gist{} (ours)}
  \end{axis}
  \end{tikzpicture}
  \vspace{4pt}
  \caption{BIG-Bench OOD accuracy (\%) vs.\ shot count $K$ on Qwen3-4B (\gist{} vs.\ few-shot ICL). The shrinking \gist{} lead with more shots indicates parametric induction gains rather than richer context.}
  \label{fig:fewshot_util}
\end{figure}
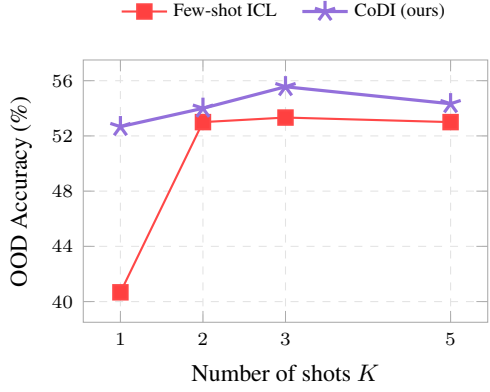

\paragraph{Reasoning turns rule-directed.}
OOD gains co-occur with shorter, rule-directed traces rather than longer CoT.
Unlike the main protocol (Table~\ref{tab:main_results}), mechanistic analysis uses thinking-enabled ICL on a trace-paired matched OOD subset, labeling \texttt{<think>} spans with E/D/P/R units (Appendix~\ref{app:think_trace_section}).
On this subset, OOD accuracy rises 104\% while trace length falls 21\%; example-recitation units fall 51\% and D$\to$P transitions rise 42\% (Figure~\ref{fig:cognitive_trace_shift}), indicating demonstrations anchor rule formation rather than copied text.
Varying $K$ on BIG-Bench OOD (Figure~\ref{fig:fewshot_util}), \gist{} leads by 12~pp at $K{=}1$ but the margin narrows with more demonstrations, so OOD gains reflect a parametric meta-inductive prior rather than richer contexts alone.

% Figure 5: grouped lollipop plot of signed trace-metric changes

% figure 6

\begin{figure}[h]
  \centering
  \resizebox{0.88\columnwidth}{!}{%
    \begin{tikzpicture}[font=\small, scale=0.92]
      %% alternating band fills (outer→inner)
      \foreach \k/\fc in {5/gray!5, 4/white, 3/gray!5, 2/white, 1/gray!5}{
        \pgfmathsetmacro{\R}{0.4*\k}
        \fill[\fc]
          (0,\R) -- ({\R*0.866},{\R*0.5}) -- ({\R*0.866},{-\R*0.5})
          -- (0,-\R) -- ({-\R*0.866},{-\R*0.5}) -- ({-\R*0.866},{\R*0.5}) -- cycle;
      }
      %% grid: 5 concentric hexagons
      \foreach \k in {1,...,5}{
        \pgfmathsetmacro{\R}{0.4*\k}
        \draw[gray!28, line width=0.4pt]
          (0,\R) -- ({\R*0.866},{\R*0.5}) -- ({\R*0.866},{-\R*0.5})
          -- (0,-\R) -- ({-\R*0.866},{-\R*0.5}) -- ({-\R*0.866},{\R*0.5}) -- cycle;
      }
      %% 6 radial axes
      \foreach \ang in {90,30,-30,-90,-150,150}{
        \pgfmathsetmacro{\ex}{2.0*cos(\ang)}
        \pgfmathsetmacro{\ey}{2.0*sin(\ang)}
        \draw[gray!38, line width=0.45pt] (0,0) -- (\ex,\ey);
      }
      %% scale labels (raw 0–5) on left side of vertical axis
      \foreach \v/\r in {1/0.4, 2/0.8, 3/1.2, 4/1.6, 5/2.0}{
        \node[gray!58, anchor=east, inner sep=1pt, font=\tiny] at (-0.07,\r) {\v};
      }
      %% axis labels: uniform single-word style
      \node[font=\footnotesize\bfseries, above, align=center, text=black!80]
        at (0, 2.28) {Compositional};
      \node[font=\footnotesize\bfseries, right, text=black!80]
        at (1.94, 1.08) {Analogical};
      \node[font=\footnotesize\bfseries, right, text=black!80]
        at (1.94,-1.08) {Pattern};
      \node[font=\footnotesize\bfseries, below, align=center, text=black!80]
        at (0,  -2.28) {Procedural};
      \node[font=\footnotesize\bfseries, left, align=right, text=black!80]
        at (-1.94,-1.08) {Constraint};
      \node[font=\footnotesize\bfseries, left, align=right, text=black!80]
        at (-1.94, 1.08) {Schema};
      %% Pre-GIST polygon
      %% scores: Comp=1.23, Anal=1.22, Patt=1.09, Proc=1.86, Cons=1.96, Sche=2.47
      \draw[NavyBlue!80, line width=1.2pt, dashed, fill=NavyBlue!65, fill opacity=0.12]
        (0,       0.492) --   % Comp  90°
        ( 0.423,  0.244) --   % Anal  30°
        ( 0.378, -0.218) --   % Patt -30°
        (0,      -0.744) --   % Proc -90°
        (-0.679, -0.392) --   % Cons -150°
        (-0.856,  0.494) -- cycle; % Sche 150°
      %% Pre-GIST vertex dots
      \foreach \x/\y in {
          0/0.492, 0.423/0.244, 0.378/-0.218,
          0/-0.744, -0.679/-0.392, -0.856/0.494}{
        \fill[NavyBlue!80] (\x,\y) circle (1.8pt);
        \fill[white]       (\x,\y) circle (0.8pt);
      }
      %% Post-GIST polygon
      %% scores: Comp=1.51, Anal=0.59, Patt=0.47, Proc=2.93, Cons=3.46, Sche=3.19
      \draw[MediumPurple, line width=1.8pt, fill=MediumPurple, fill opacity=0.20]
        (0,       0.604) --   % Comp  90°
        ( 0.204,  0.118) --   % Anal  30°
        ( 0.163, -0.094) --   % Patt -30°
        (0,      -1.172) --   % Proc -90°
        (-1.199, -0.692) --   % Cons -150°
        (-1.105,  0.638) -- cycle; % Sche 150°
      %% Post-GIST vertex dots
      \foreach \x/\y in {
          0/0.604, 0.204/0.118, 0.163/-0.094,
          0/-1.172, -1.199/-0.692, -1.105/0.638}{
        \fill[MediumPurple] (\x,\y) circle (2.2pt);
        \fill[white]        (\x,\y) circle (0.9pt);
      }
      %% Post-GIST score labels only (6 numbers, placed along each axis direction)
      %% Comp (90°): label above vertex
      \node[MediumPurple, font=\scriptsize\bfseries, above, inner sep=2pt]
        at (0, 0.80) {1.51};
      %% Anal (30°): small value—label further along axis at r≈0.60
      \node[MediumPurple, font=\scriptsize\bfseries, right, inner sep=3pt]
        at (0.52, 0.30) {0.59};
      %% Patt (−30°): small value—label further along axis at r≈0.60
      \node[MediumPurple, font=\scriptsize\bfseries, right, inner sep=3pt]
        at (0.52,-0.30) {0.47};
      %% Proc (−90°): label below vertex
      \node[MediumPurple, font=\scriptsize\bfseries, below, inner sep=2pt]
        at (0, -1.36) {2.93};
      %% Cons (−150°): label to the left of vertex
      \node[MediumPurple, font=\scriptsize\bfseries, left, inner sep=3pt]
        at (-1.42,-0.82) {3.46};
      %% Sche (150°): label to the left of vertex
      \node[MediumPurple, font=\scriptsize\bfseries, left, inner sep=3pt]
        at (-1.30, 0.75) {3.19};
      %% center dot
      \fill[gray!45] (0,0) circle (1.5pt);
      %% legend
      \node[anchor=north, inner sep=0pt] at (0,-3.08) {%
        \begin{tabular}{@{}c@{~}l@{\quad}c@{~}l@{}}
          \tikz[baseline=-0.6ex]{%
            \fill[NavyBlue!65, opacity=0.18] (0,-0.06) rectangle (0.30,0.10);
            \draw[NavyBlue!80, dashed, line width=1.2pt] (0,0.02)--(0.30,0.02);}
          & \small Pre-\gist{} &
          \tikz[baseline=-0.6ex]{%
            \fill[MediumPurple, opacity=0.22] (0,-0.06) rectangle (0.30,0.10);
            \draw[MediumPurple, line width=1.8pt] (0,0.02)--(0.30,0.02);}
          & \small Post-\gist{}
        \end{tabular}%
      };
    \end{tikzpicture}%
  }
  \caption{Radar chart of Pre- and Post-\gist{} strategy scores on six induction modes (Gemini-3-flash).}
  \label{fig:category_radar}
\end{figure}
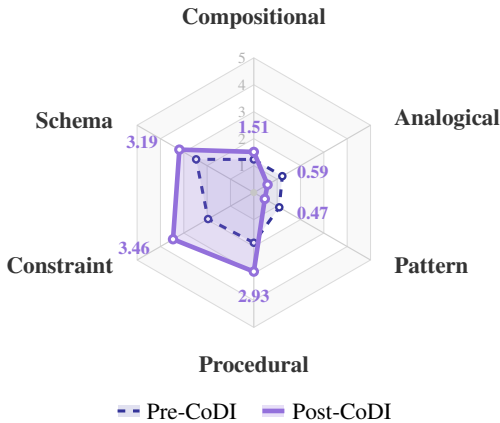

\paragraph{Strategies favor rule-based modes.}
We further find that performance gains concentrate on rule-oriented induction modes rather than extending uniformly across all strategy types.
Gemini-3-flash scores Pre- and Post-\gist{} strategies on six induction dimensions (Appendix~\ref{app:six_modes_prompt}).
Post-\gist{} improves on Constraint, Procedural and Schema but declines on Analogical and Pattern (Figure~\ref{fig:category_radar}).
This asymmetric split is consistent with classic distinctions between rule- and schema-based induction and analogical mapping~\citep{holland1986induction,gentner1983structure}; we treat pattern extrapolation as an operational category, and CoDI's execution feedback further biases the inducer toward explicit procedures.
Accordingly, the Analogical and Pattern declines mark a disclosed capability boundary rather than a uniform claim of meta-inductive superiority across all reasoning types.
Taken together with the trace analysis, the shifts support rule-driven over example-recitation induction.

%  six induction modes radar (real data)

\paragraph{Case study.}
\label{sec:induced_strategies}
Table~\ref{tab:qualitative_cases} instantiates the trace and induction-mode shifts above on one BIG-Bench Navigate episode with $K{=}3$ supports and a test query unseen in the supports.
FS treats the query as most similar to Ex3 and reuses its Yes label, which aligns with the example-recitation pattern in Figure~\ref{fig:cognitive_trace_shift} and with surface matching in \S\ref{sec:intro}.
S-FS induces a strategy without seeing the query, yet its rule restates the Ex3 trajectory and yields the same wrong Yes answer, so query-blind decoupling alone does not yield a task-family rule.
\codi{} abstracts coordinate updates verified on all three supports and executes them to reach $(2,2)$, which yields the correct No answer and aligns with the Procedural and Constraint gains in Figure~\ref{fig:category_radar}.
Accordingly, this case supports the hypothesis that generalization to unseen instances depends on rules verified across all supports rather than on query decoupling or surface matching alone.

\begin{table*}[h]
  \centering
  \footnotesize
  \setlength{\tabcolsep}{2pt}
  \renewcommand{\arraystretch}{0.98}
  \definecolor{qcfew}{RGB}{245,245,250}
  \definecolor{qctest}{RGB}{228,238,252}
  \newcommand{\qbeh}[2]{\makecell[c]{#1\\[-2pt]#2}}
  \begin{tabular}{@{}>{\centering\arraybackslash}p{0.048\textwidth} p{0.792\textwidth} @{\hspace{0.012\textwidth}} >{\centering\arraybackslash}p{0.038\textwidth} >{\centering\arraybackslash}p{0.088\textwidth}@{}}
    \toprule
    \multicolumn{4}{@{}c@{}}{%
      \begin{tabular}{@{}>{\columncolor{qcfew}}p{0.485\textwidth}|>{\columncolor{qctest}}p{0.485\textwidth}@{}}
        \centering\arraybackslash
        \textbf{Few-shot ($K{=}3$).}\ BIG-Bench Navigate; decide return to origin.\newline
        Ex1 \texttt{N2,S2}$\to$Yes;\ Ex2 \texttt{E3,W3}$\to$Yes;\newline
        Ex3 \texttt{N1,S1,E2,W2}$\to$Yes (4 moves, returns). &
        \centering\arraybackslash
        \textbf{Test query.}\ \texttt{N4,S2,E3,W1}.\newline
        Move combination unseen in supports; return to start?\newline
        Gold No (final $(2,2)\neq(0,0)$). \\
      \end{tabular}} \\
    \midrule
    \textbf{Method} &
    \multicolumn{1}{>{\centering\arraybackslash}p{0.792\textwidth}}{\textbf{Solving process}} &
    \textbf{Pred.} &
    \textbf{Behavior} \\
    \midrule
    FS &
    Compare query with all supports in context.\ Select Ex3 as the nearest match (four moves).\ Reuse the Yes label from Ex3 without coordinate tracking.\ Apply an action-count shortcut rather than origin logic, never checking whether the final position returns to $(0,0)$. &
    Yes \xmark & \qbeh{surface}{matching} \\
    S-FS &
    Induce a strategy from supports only without seeing the query.\ Rule text states track N/S/E/W and test the origin.\ At solve time, pattern-matches the test query to Ex3 and copies that move sequence instead of applying the rule.\ Output Yes without independent coordinate calculation or endpoint verification. &
    Yes \xmark & \qbeh{example}{recitation} \\
    \codi{} &
    SG induces a coordinate rule verified on Ex1--Ex3.\ AS initializes $(0,0)$ and parses four instructions.\ Updates coordinates to $(0,4)$, $(0,2)$, $(3,2)$, and $(2,2)$.\ Tests $(2,2)\neq(0,0)$ and outputs No. &
    No \cmark & \qbeh{task-level}{regularity} \\
    \bottomrule
  \end{tabular}
  \caption{%
    Navigate case study on a matched BIG-Bench Navigate episode ($K{=}3$; untrained FS/S-FS, trained \codi{}, Qwen3-4B).
    Rows compare solver processes on the same episode.
    \xmark/\cmark mark incorrect and correct predictions.%
  }
  \label{tab:qualitative_cases}
\end{table*}

% ============================================================
\subsection{What Do Induced Strategies Achieve?}
\label{sec:strategy_quality}

% \noindent These analyses test whether label-free RL produces transferable TIR that improves fixed-solver OOD execution, as claimed for \codi{} in \S\ref{sec:intro}.

\begin{figure}[h]
  \centering
  \resizebox{0.98\columnwidth}{!}{%
  \begin{tikzpicture}[x=0.075cm,y=0.70cm]
    \tikzset{
      qseg/.style={draw=black, line width=0.45pt},
      qlabel/.style={font=\small, anchor=east},
      qpct/.style={font=\small},
      qtiny/.style={font=\scriptsize, fill=white, inner sep=1pt}
    }

    % Legend
    \node[qseg, fill=qualityQwen, minimum width=0.45cm, minimum height=0.17cm] at (14,4.30) {};
    \node[font=\scriptsize, anchor=west] at (18,4.30) {Qwen3-4B Wins};
    \node[qseg, fill=qualityTie, minimum width=0.45cm, minimum height=0.17cm] at (50,4.30) {};
    \node[font=\scriptsize, anchor=west] at (54,4.30) {Tie};
    \node[qseg, fill=qualityGPT, minimum width=0.45cm, minimum height=0.17cm] at (70,4.30) {};
    \node[font=\scriptsize, anchor=west] at (74,4.30) {GPT-5 Wins};

    % Correctness: Qwen 52, Tie 2, GPT-5 46
    \node[qlabel] at (-3,3.10) {Correct.};
    \draw[qseg, fill=qualityQwen] (0,2.78) rectangle (52,3.42);
    \draw[qseg, fill=qualityTie] (52,2.78) rectangle (54,3.42);
    \draw[qseg, fill=qualityGPT] (54,2.78) rectangle (100,3.42);
    \node[qpct] at (26,3.10) {52\%};
    \node[qtiny] at (53,3.68) {2\%};
    \draw[thin] (53,3.42) -- (53,3.57);
    \node[qpct] at (77,3.10) {46\%};

    % Executability: Qwen 40, Tie 3, GPT-5 57
    \node[qlabel] at (-3,2.10) {Execut.};
    \draw[qseg, fill=qualityQwen] (0,1.78) rectangle (40,2.42);
    \draw[qseg, fill=qualityTie] (40,1.78) rectangle (43,2.42);
    \draw[qseg, fill=qualityGPT] (43,1.78) rectangle (100,2.42);
    \node[qpct] at (20,2.10) {40\%};
    \node[qtiny] at (41.5,2.68) {3\%};
    \draw[thin] (41.5,2.42) -- (41.5,2.57);
    \node[qpct] at (71.5,2.10) {57\%};

    % Transferability: Qwen 49, Tie 20, GPT-5 31
    \node[qlabel] at (-3,1.10) {Transfer.};
    \draw[qseg, fill=qualityQwen] (0,0.78) rectangle (49,1.42);
    \draw[qseg, fill=qualityTie] (49,0.78) rectangle (69,1.42);
    \draw[qseg, fill=qualityGPT] (69,0.78) rectangle (100,1.42);
    \node[qpct] at (24.5,1.10) {49\%};
    \node[qpct] at (59,1.10) {20\%};
    \node[qpct] at (84.5,1.10) {31\%};

    % Conciseness: Qwen 31, Tie 30, GPT-5 39
    \node[qlabel] at (-3,0.10) {Concise.};
    \draw[qseg, fill=qualityQwen] (0,-0.22) rectangle (31,0.42);
    \draw[qseg, fill=qualityTie] (31,-0.22) rectangle (61,0.42);
    \draw[qseg, fill=qualityGPT] (61,-0.22) rectangle (100,0.42);
    \node[qpct] at (15.5,0.10) {31\%};
    \node[qpct] at (46,0.10) {30\%};
    \node[qpct] at (80.5,0.10) {39\%};

    % Axis
    \draw[black, line width=0.4pt] (0,-0.75) -- (100,-0.75);
    \foreach \x/\lab in {0/0,20/20,40/40,60/60,80/80,100/100} {
      \draw[black, line width=0.35pt] (\x,-0.75) -- (\x,-0.88);
      \node[font=\scriptsize, anchor=north] at (\x,-0.92) {\lab};
    }
    \node[font=\scriptsize] at (50,-1.65) {Share of GPT-5 pairwise judgments (\%)};
  \end{tikzpicture}%
  }
  \caption{Pairwise GPT-5 judgments favor \codi{} on Correctness and Transferability but not on Executability or Conciseness, defining the polish-oriented boundary of RL-optimized TIR.}
  \label{fig:strategy_quality}
\end{figure}
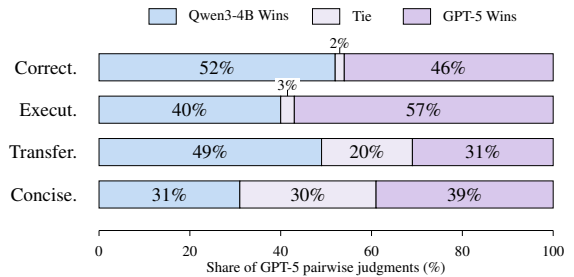

\paragraph{Correctness and transferability lead.}
We assess induced strategy quality by comparing \codi{}-Qwen3-4B and GPT-5 strategies via pairwise GPT-5 judgments on four dimensions (Appendix~\ref{app:strategy_quality_prompt}; Figure~\ref{fig:strategy_quality}).
\codi{} is preferred on Correctness and especially Transferability, indicating that label-free RL yields portable, execution-grounded TIR.
Under the same objective, Executability and Conciseness are less favored because \codi{} optimizes transferable rules for a fixed 4B executor rather than GPT-5-native phrasing, and GPT-5 self-judging may amplify this asymmetry (\S\ref{sec:limitations}).

% strategy quality

\paragraph{Gains transfer to OOD execution.}
Analysis reveals that induced strategies raise OOD accuracy under a fixed solver, with the largest margins on rule-transfer-critical task types and limited gains where baselines already saturate.
We hold Qwen3-4B \as{} fixed and vary only strategy source across five OOD types (Table~\ref{tab:strategy_source}).
\codi{} achieves the largest margin on Ctx.Param.Conf.\ (+15.0 pp over the best closed-source baseline).
On Elem.Math, Track.Shuf., and Bool.Expr., gains are modest because strong closed-source FS baselines already approach ceiling performance.
These patterns nevertheless show that label-free induced TIR improves a fixed 4B executor most where portable rules are essential, confirming transfer beyond training tasks.
Details appear in Appendix~\ref{app:solver_transfer}.

\begin{table}[t]
  \centering
  \fontsize{7.5pt}{9.5pt}\selectfont
  \setlength{\tabcolsep}{5pt}
  \renewcommand{\arraystretch}{0.98}
  \resizebox{\columnwidth}{!}{%
  \begin{tabular}{@{}l ccc@{}}
    \toprule
    \textbf{Task Type} &
    \makecell{\textbf{Gem.-Flash}\\{\footnotesize FS\,/\,S-FS}} &
    \makecell{\textbf{GPT-5}\\{\footnotesize FS\,/\,S-FS}} &
    \makecell{\textbf{Qw3-4B}\\{\footnotesize (\codi{})}} \\
    \midrule
    Ctx.Param.Conf. & 75.00\,/\,65.00  & 70.00\,/\,65.00  & \textbf{90.00} \\
    VitaminC        & 85.00\,/\,80.00  & 65.00\,/\,40.00  & \textbf{78.33} \\
    Elem.Math       & 99.00\,/\,89.00  & 95.00\,/\,75.00  & \textbf{97.67} \\
    Track.Shuf.     & 100.00\,/\,91.67 & 85.00\,/\,90.00  & \textbf{99.44} \\
    Bool.Expr.      & 90.00\,/\,95.00  & 100.00\,/\,85.00 & \textbf{98.33} \\
    \bottomrule
  \end{tabular}%
  }
  \caption{\codi{}-induced strategies yield the largest OOD gains on rule-transfer-critical types under a fixed Qwen3-4B \as{}.}
  \label{tab:strategy_source}
\end{table}

\FloatBarrier
% ============================================================
\section{Related Work}
\label{sec:related}

\paragraph{Meta-learning and in-context adaptation.}
MAML~\citep{finn2017maml} adapts parameters via gradient updates without explicit cross-instance strategies.
MetaICL~\citep{min2022metaicl} likewise meta-trains in-context learning at the parameter level.
Mirage~\citep{li2025mirage} shows that ICL often follows neighbor-based matching rather than recovering reusable abstract rules.
Motivated by these limitations, our work introduces a Cognitive-Decoupling Framework for discrete, query-blind TIR, decoupling induction from frozen execution.

\paragraph{Induction from examples and prompt optimization.}
StrategyLLM~\citep{gao2024neurips-strategyllm} induces and selects task-level strategies, while Self-Discover~\citep{zhou2024selfdiscover} composes reasoning structures for prompted tasks.
Induct-Learn~\citep{chen2024inductlearn} and LEAP~\citep{zhang2024leap} derive instructions or principles from few-shot supports, but remain inference-time prompting methods rather than a trained cross-task inducer.
CoT~\citep{wei2022cot}, STaR~\citep{zelikman2022star}, and DeepSeek-R1~\citep{deepseek2025r1} improve reasoning traces or reasoning policies; GEPA~\citep{agrawal2026gepareflectivepromptevolution} optimizes prompts via reflective search at the task level.
Our work builds upon this broader paradigm with Dual-Loop Meta-Inductive Learning, optimizing query-blind TIR via cross-task RL and EIR$\times$ICR feedback for reasoning episodes.

\paragraph{ICL under distribution shift.}
Standard ICL is highly sensitive to demonstration choice, format, label space, and input distribution, which can degrade under task-family or domain mismatch~\citep{min2022rethinking,schoch2025data,dewynter2025icl}.
Inference-time remedies such as CCL~\citep{byun2025ccl} improve OOD robustness via domain-invariant demonstration selection but remain per-query interventions rather than training a reusable inducer.
Mechanistic and theoretical work further suggests that OOD generalization in ICL depends strongly on task-distribution coverage, low-test-error preferences, and subspace alignment~\citep{wang2024incontextlearningreallygeneralize,goddard2025task,kwon2026subspace}.
In contrast, CoDI trains a query-blind inducer to learn transferable TIR from supports via cross-task RL, offering a reusable alternative to inference-time demonstration engineering for OOD few-shot generalization.

% ============================================================
\section{Conclusion}
\label{sec:conclusion}

\noindent Few-shot OOD generalization demands explicit, optimizable task-level induction beyond execution scale.
To this end, we presented \codi{}, combining a Cognitive-Decoupling Framework for query-blind TIR with Dual-Loop Meta-Inductive Learning for label-free cross-task inducer optimization.
Experiments show that neither parameter-level adaptation nor query-blind decoupling alone yields reusable TIR, yet TIR-aligned training drives systematic OOD gains across benchmarks.
Accordingly, meta-induction constitutes a trainable capability complementary to model scale.
Future work may evaluate finer task-level shifts, strengthen analogical and pattern induction, and study strategy--solver co-adaptation across languages and modalities.

% ============================================================
\section*{Limitations}
\label{sec:limitations}

\paragraph{Computational cost.}
Each training episode requires $G \times M = 32$ frozen \as{} forward passes, roughly $32\times$ standard SFT compute.
Training completes in 12--24 hours on 8$\times$A800 GPUs; future work may reduce cost via adaptive rollout schedules, cached solver verification, or distillation from the trained \sg{}.

\paragraph{Benchmark scope and confounds.}
We evaluate only on English text-reasoning benchmarks, leaving broader language, modality, and distribution shifts untested.
Gemini-3-flash both proposes SFT strategies and annotates mechanistic traces, and ID/OOD splits share the StrategyBench construction pipeline, but GPT-5 cross-annotation (Appendix~\ref{app:crosscheck}) and gains on MATH-500, StrategyQA, and ReClor partially address these concerns.

\paragraph{Strategy--solver coupling.}
We fix \as{} during training and update \sg{} via execution feedback for stable credit assignment.
TIR is natural-language and interface-decoupled, yet effective quality depends on the training-time executor.
Under a fixed executor (Table~\ref{tab:strategy_source}), substitution gains favor rule-governed over analogical or pattern modes and support portable structure, while strategy--solver co-adaptation remains out of scope.

% ============================================================
\section*{Acknowledgments}

\noindent This work is supported by the AI9Stars community, the Foundation of Muyuan Laboratory (Program ID: 11116022401), and the National Natural Science Foundation of China (Grant No.\ 32370095).
Chaojun Xiao is supported by the China National Postdoctoral Program for Innovative Talents (grant no.\ BX20250388).

% ============================================================
\bibliography{references}

% ============================================================
\appendix
\raggedbottom
\setlength{\textfloatsep}{8pt plus 2pt minus 4pt}
\setlength{\floatsep}{8pt plus 2pt minus 4pt}
\setlength{\intextsep}{8pt plus 2pt minus 4pt}

\section{Implementation Details}
\label{app:training}

\noindent Pretrained backbones and source datasets follow their public licenses (Qwen3, Llama~3.1, Gemma~3, and BIG-Bench).
CoDI training code is linked in the abstract and released under the MIT License.
StrategyBench construction scripts and task-level splits are given in Appendix~\ref{app:strategybench}.

\subsection{Models, Evaluation Protocols, and Baselines}
\label{app:models}

\noindent \textbf{Models and APIs.}
We train and evaluate \codi{} on \textbf{Qwen3-4B}~\citep{yang2025qwen3technicalreport}, \textbf{Llama-3.1-8B}~\citep{dubey2024llama3}, and \textbf{Gemma-3-4B}~\citep{gemmateam2025gemma3}.
Qwen3-8B uses the same recipe and is omitted from the main table for space.
Closed-source APIs (\textbf{Gemini-3-flash}~\citep{google2025gemini3flash} and \textbf{GPT-5}~\citep{openai2025gpt5}) are inference-time upper bounds and are not fine-tuned.
\textbf{Gemini-3-flash} denotes Google AI model ID \texttt{gemini-3-flash-preview}.
It proposes SFT reference strategies and serves as the primary annotator in mechanistic analysis.
It is not used for task filtering or ID/OOD split construction.
Task filtering, categorization, and split assignment use \textbf{Qwen3-32B} scoring with author spot-checks (Appendix~\ref{app:strategybench}).
\textbf{Gemma-3-4B} denotes \texttt{gemma-3-4b-it}~\citep{google2025gemma34bit} and \textbf{Llama-3.1-8B} denotes \texttt{Llama-3.1-8B-Instruct}~\citep{meta2024llama318binstruct}.

\textbf{Baselines and implementation.}
MetaICL and GEPA are described in \S\ref{sec:setup}.
GEPA uses \texttt{gemini-3-flash-preview} as the optimizer and Qwen3-4B as the executor, with four reflective iterations.
Defaults are $G{=}8$, $K{=}3$, and $\beta{=}0.5$ on 8$\times$A800 GPUs; full hyperparameters are in Table~\ref{tab:hyperparams} and the training pipeline below.

\textbf{Numeric format.}
Reported accuracies are percentages (e.g., 45.00 denotes 45.00\%); $\Delta$\% is the relative change from the same-backbone FS baseline (\S\ref{sec:setup}).

\subsection{Hyperparameters}

\noindent Table~\ref{tab:hyperparams} lists the default training configuration used in all main experiments.

\begin{table}[h]
  \centering
  \small
  \setlength{\tabcolsep}{5pt}
  \renewcommand{\arraystretch}{1.12}
  \resizebox{\columnwidth}{!}{%
  \begin{tabular}{ll}
    \toprule
    \textbf{Hyperparameter} & \textbf{Value} \\
    \midrule
    \multicolumn{2}{l}{\textit{Model configuration}} \\
    Strategy Generator (\sg{}) & Qwen3-4B (trainable) \\
    Answer Solver (\as{})       & Qwen3-4B (frozen, non-thinking mode) \\
    SFT demonstrations          & Gemini-3-flash proposals, execution-filtered \\
    Task filtering / splits     & Qwen3-32B scoring + author spot-checks \\
    Hardware                    & 8$\times$A800 (80\,GB) \\
    Training time               & ${\sim}$12--24 h (one full \codi{} run) \\
    \midrule
    \multicolumn{2}{l}{\textit{Training hyperparameters}} \\
    Rollout count $G$             & 8 \\
    Few-shot shots $K$            & 3 \\
    Learning rate                 & $1\times10^{-6}$ (AdamW) \\
    Max generation length         & 16{,}384 tokens \\
    Training data                 & ${\sim}$20K samples \\
    \midrule
    \multicolumn{2}{l}{\textit{Reward parameters}} \\
    Asymm.\ weight $\alpha^+$ (positive gain) & 0.7 \\
    Asymm.\ weight $\alpha^-$ (negative gain) & 0.3 \\
    ICR amplification $\beta$     & 0.5 \\
    Log-smoothing constant $k$    & 10 \\
    Lower-bound cutoff $\varepsilon$  & $0.05$ \\
    Baseline EMA decay $\gamma$   & 0.9 \\
    \bottomrule
  \end{tabular}%
  }
  \caption{Hyperparameters for \codi{} training.}
  \label{tab:hyperparams}
\end{table}

\paragraph{Two-stage pipeline.}
\codi{} training comprises Phase~1 (SFT warm-up) and Phase~2 (inductive alignment).

\paragraph{Phase 1 SFT warm-up.}
$\pi_{\sg{}}$ is fine-tuned on teacher-constructed demonstrations to emit format-compliant strategies.
Supervision targets schema validity rather than task accuracy.
After a few hundred steps (${>}90\%$ format-valid rate), $\pi_{\sg{}}^{(0)}$ initializes per-task EIR baselines $\{b_\tau\}$ from inference accuracies on training tasks.

\paragraph{Phase 2 inductive alignment.}
We optimize $\pi_{\sg{}}$ with the joint reward in Eq.~\eqref{eq:total_reward} via cross-task GRPO while \as{} remains frozen.
Format-invalid rollouts are excluded from \as{} execution and receive a reward of zero, as specified in the format gate below.
Each episode samples one task type uniformly at random.

\paragraph{Format gating.}
A binary check tests whether all required schema sections are present and structurally valid.
Format-invalid strategies receive $R_i = 0$ and are not passed to \as{}, so EIR and ICR are not computed.
A zero reward, rather than a negative penalty, avoids inflating reward variance while withholding credit for malformed outputs, which lets GRPO move away from format violations without destabilizing training.

\subsection{EIR: Difficulty-Adaptive Normalizer}
\label{app:eir_normalizer}

\noindent The normalizer $c(b_\tau)$ in Eq.~\eqref{eq:eir_raw} is defined as
\begin{equation}
  c(b_\tau) = \max\!\bigl(b_\tau^{\,\alpha}(1-b_\tau)^{1-\alpha},\; \varepsilon\bigr)
  \label{eq:c_btau}
\end{equation}
where $\alpha$ is selected by the sign of $\delta_i$ and $\varepsilon = 0.05$ lower-bounds the denominator when $b_\tau$ approaches $0$ or $1$.

\paragraph{Mathematical properties.}
The function $b^\alpha(1-b)^{1-\alpha}$ on $b \in [0,1]$ is an unnormalized Beta$(\alpha{+}1, 2{-}\alpha)$ density, maximized uniquely at $b=\alpha$.
When $b_\tau=\alpha$, $c(b_\tau)$ is largest and $|g_\text{raw}|$ is smallest; as $b_\tau$ moves toward easy ($b_\tau \to 1$) or hard ($b_\tau \to 0$) extremes, $c(b_\tau)$ shrinks and $|g_\text{raw}|$ grows, up-weighting deviations on extreme-difficulty tasks.

\paragraph{Role of asymmetric $\alpha$.}
With $\alpha^+ = 0.7 > \alpha^- = 0.3$, EIR applies two difficulty calibrations:
\begin{itemize}[leftmargin=*, nosep]
  \item \textbf{Positive ($\delta_i \geq 0$, $\alpha = \alpha^+$):} $c(b_\tau)$ peaks at $b_\tau = 0.7$, giving mid-to-high difficulty tasks a natural baseline while amplifying gains on very hard tasks ($b_\tau \ll 0.7$).
  \item \textbf{Negative ($\delta_i < 0$, $\alpha = \alpha^-$):} $c(b_\tau)$ peaks at $b_\tau = 0.3$; regressions on high-baseline tasks ($b_\tau \gg 0.3$) receive stronger penalties, whereas low-baseline regressions retain larger $c(b_\tau)$ and preserve early exploration.
\end{itemize}

\paragraph{Ablation.}
Setting $\alpha^+ = \alpha^- = 0.5$ causes near-complete training collapse (BBH 4.26, OOD 3.67; Table~\ref{tab:hparam_study}), confirming that asymmetric EIR weighting is necessary for stable cross-task GRPO.

\section{Full Strategy Schema and Format Comparison}
\label{app:schema}

\noindent This section compares three strategy formats, specifies the \codi{} semi-structured template, and gives two cross-task examples.
Ablation numbers are in Table~\ref{tab:ablation_format}.

\subsection{Format comparison, specification, and examples}
\label{app:three_formats}

\noindent We use BIG-Bench Navigate (grid tracking, decide whether the agent returns to the origin) to compare Free-form, CoT-style, and Semi-structured (\codi{}) formats, matching Table~\ref{tab:ablation_format}.
The table also ablates the mandatory sections \texttt{[INDUCTIVE\_RULE]}, \texttt{[TASK\_PROCEDURE]}, and \texttt{[TRANSFER\_SCOPE]}.

\paragraph{Free-form (unstructured).}
The inducer is prompted as follows.
\begin{quote}\small\ttfamily
Based on the following examples, generate a strategy
to solve this type of problem.

[few-shot examples]

Strategy:
\end{quote}
A typical \sg{} output is as follows.
\begin{quote}\small\ttfamily
Track coordinates. North +y, South -y, East +x, West -x.
Check if (x,y) = (0,0) at end.
\end{quote}
\noindent\textbf{Limitation:} Output is unconstrained in length and structure, so token-level log-probability baselines are incomparable across rollouts and ICR is ill-defined.
\sg{} often paraphrases specific examples rather than stating verifiable task-family rules, yielding the weakest OOD generalization.

\paragraph{CoT-style (implicit structure).}
The inducer is prompted as follows.
\begin{quote}\small\ttfamily
Based on the following examples, analyze the pattern
and generate a step-by-step strategy for solving this
type of problem. Think carefully and reason step by step.

[few-shot examples]

Strategy:
\end{quote}
A typical \sg{} output is as follows.
\begin{quote}\small\ttfamily
Let me analyze the examples:
- Example 1: coordinates update with each direction...
- Example 2: same pattern confirmed...

General approach:
1. Initialize position (0,0)
2. Parse each instruction...
3. Check if final position is origin

This should work for similar grid navigation problems,
though diagonal movements might need special handling.
\end{quote}
\noindent\textbf{Limitation:} Reasoning steps appear, but rules and procedures are interleaved and boundary conditions are ad hoc.
Without fixed section headers, \as{} sees variable prompts, ICR is only moderately reliable, and OOD transfer is weaker than under semi-structured control.

\paragraph{Semi-structured prompt of \codi{}:}
\begin{quote}\small\ttfamily
Based on the following examples, generate a strategy
using EXACTLY the following sections:
[INDUCTIVE\_RULE] / [TASK\_PROCEDURE] / [TRANSFER\_SCOPE]

[few-shot examples]

Strategy:
\end{quote}
\noindent\textbf{Typical \sg{} output:} 
\begin{quote}\small\ttfamily
[INDUCTIVE\_RULE]
Update coordinates: North (+y), South (-y), East (+x), West (-x).
Decide whether the endpoint equals the origin (0,0).

[TASK\_PROCEDURE]
1. Initialize (x,y)=(0,0)
2. Parse each instruction for direction and steps
3. Update coordinates
4. Output Yes if (x,y)==(0,0), else No

[TRANSFER\_SCOPE]
Applies when directional moves are given and the query asks about returning to the start.
Not applicable when diagonal moves or conditional branches appear; extend Step 2 accordingly.
\end{quote}
\noindent\textbf{Advantage:} Fixed section headers yield comparable token-level baselines and measurable ICR; \texttt{INDUCTIVE\_RULE} enforces explicit induction and \texttt{TRANSFER\_SCOPE} manages boundaries, giving the strongest OOD performance.

\paragraph{Semi-structured specification.}
\label{app:schema_spec}
\codi{} requires \sg{} to emit three named sections; we define their semantics and format violations below.

\paragraph{Mandatory sections.}
Each strategy must contain three sections delimited by header lines:

\begin{itemize}[leftmargin=*, itemsep=2pt]
  \item \textbf{\texttt{[INDUCTIVE\_RULE]}}: Task-family regularities inferred from $K$ examples, verifiable against supports but not tied to any single instance. Boundary cases belong in \texttt{[TRANSFER\_SCOPE]}.
  \item \textbf{\texttt{[TASK\_PROCEDURE]}}: Three to six numbered, executable steps covering parsing, reasoning, and output formatting.
  \item \textbf{\texttt{[TRANSFER\_SCOPE]}}: Scope specification with trigger conditions, at least one non-applicable case, and any exceptions to the inductive rule.
\end{itemize}

Format violations (missing sections, misspelled headers, or empty section bodies) assign $R_i=0$ and exclude the strategy from \as{} execution (format gating above).

\paragraph{Worked example (Navigate).}
The following Navigate strategy illustrates the three-section template.

\begin{quote}\small\ttfamily
[INDUCTIVE\_RULE]

From the examples: each instruction has the form ``walk \{steps\} steps \{direction\}.''
Update North (+y), South (-y), East (+x), West (-x); test whether the final point is (0,0).

[TASK\_PROCEDURE]

1. Initialize (x, y) = (0, 0).
2. Parse each instruction for direction and step count.
3. Update coordinates accordingly.
4. After all instructions, check whether (x, y) = (0, 0).
5. Output ``Yes'' if true, otherwise ``No.''

[TRANSFER\_SCOPE]

Trigger: a sequence of directional moves with a return-to-start query.
Application: apply the coordinate-tracking steps directly.
Non-applicable: (1) diagonal moves; (2) conditional branches (e.g., ``if ..., then go north''), which require extra parsing in Step 2.
\end{quote}

\paragraph{Worked example (Word sorting).}
The following Word Sorting example shows the same schema on a different task family.

\begin{quote}\small\ttfamily
[INDUCTIVE\_RULE]

Sort the given word list alphabetically (case-insensitive).
Compare characters left to right; digits precede letters; special characters follow ASCII order.

[TASK\_PROCEDURE]

1. Lowercase all words for comparison.
2. Sort by alphabetical order with character-wise comparison.
3. If prefixes match, place the shorter word first.
4. Output the sorted list as comma-separated tokens.

[TRANSFER\_SCOPE]

Trigger: unordered word list requiring alphabetical output.
Application: follow Steps 1--4.
Non-applicable: (1) frequency- or semantics-based sorting requires a new criterion in Step 2; (2) multi-word phrases must be treated as single tokens in Step 1.
\end{quote}

% ============================================================
\section{Annotation and Evaluation Prompts}
\label{app:prompts}

\noindent Figures~\ref{fig:six_modes_prompt}--\ref{fig:inductive_rule_eval_prompt} collect the full templates used for mechanistic scoring.
They appear at the end of the appendix in single-column layout.

\paragraph{Six-mode scoring.}
\label{app:six_modes_prompt}
Figure~\ref{fig:six_modes_prompt} is the Gemini-3-flash rubric for six induction modes (integer scores $0$--$5$).
The rubric is an operationalization inspired by classic accounts of induction and analogy~\citep{holland1986induction,gentner1983structure}.
Compositional and Procedural scores target \texttt{[TASK\_PROCEDURE]}; Analogical and Pattern Extrapolation target \texttt{[INDUCTIVE\_RULE]}; Constraint-based and Schema target \texttt{[TRANSFER\_SCOPE]}.

\paragraph{Think-unit tagging.}
Figure~\ref{fig:think_unit_prompt} tags sentence-level \texttt{<think>} units as E/D/P/R.

\paragraph{Strategy quality.}
\label{app:strategy_quality_prompt}
Figure~\ref{fig:strategy_quality_prompt} is the GPT-5 pairwise judge for Correctness, Executability, Transferability, and Conciseness, with models presented as A and B.

\paragraph{Inductive-rule judging.}
Figure~\ref{fig:inductive_rule_eval_prompt} scores each \texttt{[INDUCTIVE\_RULE]} on abstraction, coverage, copying, and transferability.
Aggregate judge means are not reported; the template is included so the LOO qualitative protocol can be reproduced.

% ============================================================
\section{Thinking-Unit Taxonomy and Trace Metrics}
\label{app:think_trace_section}

\noindent This section supports \S\ref{sec:reasoning_process} and Figure~\ref{fig:cognitive_trace_shift}.
Unlike Table~\ref{tab:main_results}, which uses strategy-then-answer with thinking disabled, this analysis uses thinking-enabled direct ICL on a trace-paired matched OOD subset (episodes where both Pre and Post checkpoints emit a valid \texttt{<think>} span).
We split each \texttt{<think>} span into sentence-level units and assign E/D/P/R labels with Gemini-3-flash (Figure~\ref{fig:think_unit_prompt}).

\subsection{Unit taxonomy (E/D/P/R)}
\label{app:think_unit_labels}

\noindent Table~\ref{tab:think_unit_taxonomy} defines the four sentence-level tags used in this section.
E and D are non-inductive. P marks a newly stated rule. R marks application of a rule already stated in the same trace.

\begin{table}[h]
  \centering
  \small
  \setlength{\tabcolsep}{4pt}
  \renewcommand{\arraystretch}{1.1}
  \resizebox{\columnwidth}{!}{%
  \begin{tabular}{@{}c l c p{4.35cm}@{}}
    \toprule
    \textbf{Tag} & \textbf{Name} & \textbf{Ind.?} & \textbf{Description} \\
    \midrule
    E & Example-recitation & No &
    The unit recalls, quotes, or closely paraphrases a \emph{specific} support example without abstracting beyond it. \\
    D & Direct-application & No &
    The unit applies facts or procedures to the current instance without first stating a cross-example pattern. \\
    P & Pattern-abstraction & Yes &
    The unit states a general rule or invariant intended to hold across instances (inductive content). \\
    R & Rule-application & Yes &
    The unit applies a rule or pattern that was already stated earlier in the trace. \\
    \bottomrule
  \end{tabular}%
  }
  \caption{Cognitive tags for sentence-level units inside \texttt{<think>}. ``Ind.?'' indicates inductive (rule-level) content; descriptions define the annotation target.}
  \label{tab:think_unit_taxonomy}
\end{table}

\paragraph{Phase-1 prefix.}
Let the labeled unit sequence be $(y_1,\ldots,y_n)$.
\textbf{Phase-1} is the maximal prefix of consecutive $E$ labels at the start of the trace: indices $\{1,\ldots,t\}$ where $y_1=\cdots=y_t=E$ and either $t=n$ or $y_{t+1}\neq E$.
We report $n_E^{(\mathrm{ph1})}=\sum_{i\le t}\mathbf{1}[y_i{=}E]$, which equals $t$ on this definition.
Figure~\ref{fig:cognitive_trace_shift} plots this prefix length, not the total E count $n_E$ used in $q_{E/R}$ and $\rho_E$.

\subsection{Trace metrics and statistical tests}
\label{app:think_unit_metrics}

\noindent Let $(y_1,\ldots,y_n)$ be labels after dropping empty segments; $n$ is the number of units.
Denote counts $n_c=\sum_i \mathbf{1}[y_i{=}c]$ for $c\in\{E,D,P,R\}$, and $n_{P\cup R}=n_P+n_R$.
\textbf{Example-to-rule ratio.} Whenever $n_{P\cup R}>0$, define $q_{E/R}=n_E/n_{P\cup R}$.
\textbf{Example share.} $\rho_E=n_E/n$, reported as a percentage.
\textbf{Think length.} $L_{\mathrm{think}}$ is the character length of the full \texttt{<think>} span (including whitespace).
\textbf{D$\to$P rate.} $r_{D\to P}=\frac{1}{n-1}\sum_{i=1}^{n-1}\mathbf{1}[y_i{=}D,\,y_{i+1}{=}P]$ (percentage of adjacent pairs).
\textbf{Mean P length.} $\bar{\ell}_P=\mathrm{mean}\{\mathrm{len}(\text{unit}_i)\mid y_i{=}P\}$ (characters); undefined if $n_P{=}0$ (those samples are excluded from $\bar{\ell}_P$ averages only).
\textbf{Accuracy.} $\mathrm{Acc}_{\mathrm{OOD}}^{\mathrm{matched}}$ is exact-match accuracy on the same thinking-enabled matched subset as Fig.~\ref{fig:cognitive_trace_shift}, not the full OOD split in Table~\ref{tab:main_results}.

\paragraph{Inference and testing.}
All sentence labels are produced by Gemini-3-flash with temperature $0$.
GPT-5 re-annotation on a random subsample of 40 matched OOD tasks (180 sentence-level units) yields Cohen's $\kappa{=}0.79$ on E/D/P/R labels and Pearson $r{=}0.91$ on six-mode scores (Appendix~\ref{app:crosscheck}; Tables~\ref{tab:crosscheck_modes} and~\ref{tab:crosscheck_traces}).
Unless otherwise noted, $p$-values are two-sided paired tests at the trajectory level over matched tasks (Pre vs.\ Post).
Table~\ref{tab:think_unit_significance} reports the primary process metrics.
Matched accuracy rises from 45.2 to 92.4 while think length and the phase-1 recitation prefix fall, and D$\to$P transitions and mean P length rise.

\begin{table}[H]
  \centering
  \small
  \setlength{\tabcolsep}{4pt}
  \renewcommand{\arraystretch}{1.12}
  \begin{tabular}{@{}>{\raggedright\arraybackslash}p{0.31\columnwidth} @{\hspace{2pt}} c @{\hspace{2pt}} c @{\hspace{2pt}} c @{\hspace{2pt}} c@{}}
    \toprule
    \textbf{Metric} & \textbf{Pre} & \textbf{Post} & \textbf{$\Delta$} & \textbf{$p$} \\
    \midrule
    $L_{\mathrm{think}}$ & 5276 & 4171 & $-1105$ & $<10^{-3}$ \\
    $q_{E/R}$ & 1.079 & 0.674 & $-0.405$ & $3.0{\times}10^{-4}$ \\
    $r_{D\to P}$ (\%) & 4.36 & 6.21 & $+1.85$ & $4.0{\times}10^{-5}$ \\
    $\bar{\ell}_P$ & 89.4 & 108.6 & $+19.2$ & $8{\times}10^{-6}$ \\
    \midrule
    \makecell[l]{Acc$_{\mathrm{OOD}}$\\(matched, \%)} & 45.2 & 92.4 & $+47.2$ & $<10^{-3}$ \\
    $n_E^{(\mathrm{ph1})}$ & 18.54 & 9.06 & $-9.48$ & $2.1{\times}10^{-4}$ \\
    $\rho_E$ (\%) & 16.50 & 12.14 & $-4.36$ & $1.3{\times}10^{-3}$ \\
    \bottomrule
  \end{tabular}
  \caption{Pre/Post-\gist{} trace metrics on Qwen3-4B matched OOD tasks with valid thinking traces ($K{=}3$; subset of Table~\ref{tab:main_results} OOD).
  Acc.: McNemar; $n_E^{(\mathrm{ph1})}$, $\rho_E$: paired $t$; remaining metrics: paired trajectory tests.}
  \label{tab:think_unit_significance}
\end{table}

% ============================================================
\section{GPT-5 Cross-Annotation Check}
\label{app:crosscheck}

\noindent To test whether mechanistic conclusions reflect Gemini-3-flash annotation preferences, GPT-5 re-scores strategies and re-annotates traces on a random subsample of 40 matched OOD tasks (180 sentence-level units), using the rubrics in Appendix~\ref{app:six_modes_prompt} and Fig.~\ref{fig:think_unit_prompt}.
The six-mode rubric treats the six dimensions as distinct~\citep{holland1986induction,gentner1983structure}, so a high score on one mode does not imply a high score on another.

\paragraph{Six-mode strategy scores.}
Table~\ref{tab:crosscheck_modes} reports Pre-/Post-\gist{} scores on the six induction modes.
Gemini-3-flash columns repeat the full-cohort scores of Figure~\ref{fig:category_radar}.
GPT-5 columns are the $n{=}40$ subsample.
Rule-oriented modes (Procedural, Constraint, Schema) rise after training under both annotators, whereas Analogical and Pattern decline.
Every Post$-$Pre $\Delta$ keeps the same sign, which is a sign check rather than a claim of equal sample size.

\begin{table}[h]
  \centering
  \small
  \setlength{\tabcolsep}{3pt}
  \renewcommand{\arraystretch}{1.08}
  \begin{tabular}{@{}l c c c c@{}}
    \toprule
    \textbf{Mode} &
    \textbf{Gem.\ Pre} &
    \textbf{Gem.\ Post} &
    \textbf{G5 Pre} &
    \textbf{G5 Post} \\
    \midrule
    Compositional & 1.23 & 1.51 & 1.18 & 1.46 \\
    Procedural    & 1.86 & 2.93 & 1.81 & 2.86 \\
    Constraint    & 1.96 & 3.46 & 1.88 & 3.38 \\
    Schema        & 2.47 & 3.19 & 2.41 & 3.12 \\
    Analogical    & 1.22 & 0.59 & 1.15 & 0.54 \\
    Pattern       & 1.09 & 0.47 & 1.14 & 0.53 \\
    \bottomrule
  \end{tabular}
  \caption{Pre-/Post-\gist{} six-mode scores (Gem.: full cohort, Fig.~\ref{fig:category_radar}; G5: $n{=}40$).
  All six Post$-$Pre $\Delta$ share the same sign across annotators.}
  \label{tab:crosscheck_modes}
\end{table}

\paragraph{Trace-level Post$-$Pre deltas.}
Table~\ref{tab:crosscheck_traces} reports Post$-$Pre changes for Gemini-3-flash on the full matched cohort (Table~\ref{tab:think_unit_significance}) and for GPT-5 on the $n{=}40$ subsample (180 units).
All continuously scored metrics keep the same sign, as do the six-mode Post$-$Pre directions in Table~\ref{tab:crosscheck_modes}.
The comparison is a sign check rather than a matched-pair test on identical items.

\begin{table}[h]
  \centering
  \small
  \setlength{\tabcolsep}{5pt}
  \renewcommand{\arraystretch}{1.08}
  \begin{tabular}{@{}l c c c@{}}
    \toprule
    \textbf{Metric} & \textbf{Gem.\ $\Delta$} & \textbf{G5 $\Delta$} & \textbf{Sign} \\
    \midrule
    $L_{\mathrm{think}}$ & $-1105$ & $-968$ & \cmark \\
    $q_{E/R}$ & $-0.405$ & $-0.382$ & \cmark \\
    $r_{D\to P}$ & $+1.85$ & $+1.72$ & \cmark \\
    $\bar{\ell}_P$ & $+19.2$ & $+18.6$ & \cmark \\
    $n_E^{(\mathrm{ph1})}$ / $\rho_E$ & $-9.48$ / $-4.36$ & $-9.12$ / $-4.08$ & \cmark \\
    Six-mode signs & 6/6 & 6/6 & \cmark \\
    \bottomrule
  \end{tabular}
  \caption{Post$-$Pre trace deltas: Gem.\ (full cohort) vs.\ G5 ($n{=}40$).
  Sign agreement only (Appendix~\ref{app:think_unit_metrics}).}
  \label{tab:crosscheck_traces}
\end{table}

% ============================================================
\section{Sensitivity Analysis}
\label{app:sensitivity}

\noindent Table~\ref{tab:hparam_study} reports hyperparameter sensitivity for $K$ (shot count in the strategy-then-answer pipeline of Table~\ref{tab:main_results}), $\beta$, $\alpha$, and $k$ (log-smoothing constant in EIR) on Qwen3-4B.
This $K$ sweep is distinct from the evaluation-time shot curve in Figure~\ref{fig:fewshot_util}.
$^\star$ marks the default in main experiments.

\begin{table}[t]
  \centering
  \small
  \setlength{\tabcolsep}{4pt}
  \renewcommand{\arraystretch}{0.95}
  \begin{tabular}{@{}l c c c c@{}}
    \toprule
    \textbf{Parameter} & \textbf{Value} &
    {\textbf{ID}} & {\textbf{BBH}} & {\textbf{OOD}} \\
    \midrule
    \multirow{4}{*}{$K$ (shots)}
      & 1              & 48.89 & 44.20 & 34.71 \\
      & 2              & 55.67 & 68.52 & 48.00 \\
      & $3^{\star}$    & 64.67 & 82.22 & 63.00 \\
      & 5              & 63.67 & 78.52 & 57.67 \\
    \midrule
    \multirow{3}{*}{$\beta$ (ICR weight)}
      & 0.3            & 50.00 & 44.20 & 44.67 \\
      & $0.5^{\star}$  & 64.67 & 82.22 & 63.00 \\
      & 0.7            & 39.00 & 48.89 & 47.00 \\
    \midrule
    \multirow{3}{*}{$\alpha^-/\alpha^+$}
      & $0.1/0.9$                 & 55.67 & 71.85 & 47.33 \\
      & $0.3/0.7^{\star}$         & 64.67 & 82.22 & 63.00 \\
      & $0.5/0.5$                 &  4.00 &  4.26 &  3.67 \\
    \midrule
    \multirow{4}{*}{$k$ (log-smooth.)}
      & 1                         & 30.00 & 35.00 & 34.00 \\
      & 5                         & 37.33 & 41.85 & 36.67 \\
      & $10^{\star}$              & 64.67 & 82.22 & 63.00 \\
      & 20                        & 41.67 & 49.81 & 36.00 \\
    \bottomrule
  \end{tabular}
  \caption{Hyperparameter sensitivity on Qwen3-4B for $K$, $\beta$, $\alpha$, and $k$ ($^\star$: default; exact-match accuracy, \%).}
  \label{tab:hparam_study}
\end{table}

\paragraph{Shot count $K$.}
Performance peaks at $K{=}3$ (BBH 82.22, OOD 63.00) and drops slightly at $K{=}5$ (BBH 78.52, OOD 57.67).
The largest single-step gain is $K{=}1{\to}2$ ($+24.32$ BBH).
At $K{=}1$, OOD is 34.71, below the untrained FS baseline of 53.33 in Table~\ref{tab:main_results}, so one-shot training is insufficient for the full strategy-then-answer pipeline.

\paragraph{ICR amplification $\beta$.}
$\beta{=}0.5$ is optimal across all metrics.
Performance degrades at both $\beta{=}0.3$ (BBH 44.20, OOD 44.67) and $\beta{=}0.7$ (BBH 48.89, OOD 47.00).
Too little amplification under-weights the inductive signal, while too much distorts the reward landscape and destabilizes GRPO.

\paragraph{EIR asymmetry $\alpha$.}
Asymmetric weighting $(\alpha^+,\alpha^-)=(0.7,0.3)$ yields the best result (OOD 63.00).
The symmetric setting $(0.5,0.5)$ causes near-complete training collapse (BBH 4.26, OOD 3.67), so the asymmetric design is needed for GRPO stability across task difficulties.
A more extreme pair $(\alpha^-,\alpha^+)=(0.1,0.9)$ is worse (OOD 47.33), which suggests that over-penalizing negative outcomes suppresses exploration on hard tasks.

\paragraph{Log-smoothing constant $k$.}
$k$ controls logarithmic compression of the difficulty-normalized gain $g_{\text{raw}}$ in EIR and is distinct from shot count $K$.
The default $k{=}10$ is best across splits (BBH 82.22, OOD 63.00).
At $k{=}1$, insufficient smoothing yields noisy rewards and unstable training (BBH 35.00, OOD 34.00).
Performance rises at $k{=}5$ (BBH 41.85, OOD 36.67) and peaks at $k{=}10$.
Increasing $k$ to $20$ over-compresses difficulty gradients (BBH 49.81, OOD 36.00).
Weak or strong compression both hurt; $k{=}10$ is the main default.

% ============================================================
\section{Solver Transferability}
\label{app:solver_transfer}

\noindent Table~\ref{tab:solver_transfer} fixes \gist{}-4B \sg{} and swaps \as{} to test strategy transfer across executors.

\begin{table}[t]
  \centering
  \small
  \setlength{\tabcolsep}{5pt}
  \renewcommand{\arraystretch}{1.08}
  \begin{tabular}{@{}l c c c@{}}
    \toprule
    \textbf{\sg{}} & \textbf{\as{}} & \textbf{OOD}$\uparrow$ &
    \makecell[c]{$\bm{\Delta}$\\{vs.\ \gist{}-4B \as{}}} \\
    \midrule
    \multirow{4}{*}{\gist{}-4B}
      & 4B         & 61.93 & $-1.07$ \\
      & 8B         & 62.89 & $-0.11$ \\
      & 14B        & 51.14 & $-11.86$ \\
      & 32B        & 61.30 & $-1.70$ \\
    \bottomrule
  \end{tabular}
  \caption{Solver transfer with a fixed \gist{}-4B \sg{}.
  Deltas are absolute OOD points relative to the main-table \gist{}-4B \as{} score of 63.00.
  The 4B row is a re-run under the swap protocol and need not equal 63.00.}
  \label{tab:solver_transfer}
\end{table}

\noindent The 4B row (61.93) follows the swap protocol, not the main-table OOD (63.00).
4B, 8B, and 32B \as{} remain close to that reference (61.93, 62.89, and 61.30), whereas 14B drops to 51.14.
The 14B dip may reflect instruction-tuning mismatch.
Solver compatibility is left to future work.

% ============================================================
\section{LOO Rule Sensitivity and Inductive-Rule LLM Judging}
\label{app:loo_rule_diagnostics}

\noindent Figure~\ref{fig:induction_mechanism} reports embedding-based leave-one-out (LOO)
sensitivity of the \texttt{[INDUCTIVE\_RULE]} span on matched evaluation tasks
with $K{=}3$.
The four-dimension GPT-5 rubric is given in Fig.~\ref{fig:inductive_rule_eval_prompt} for reproducibility; we do not report aggregate judge means.
Table~\ref{tab:fig4_task_index} maps heatmap rows T01--T10 to task types and splits.
T01 and T05 are ID. T02 and T07 are OOD-subtask (Table~\ref{tab:id_ood_tasks}).
T10 (Date understanding) is a BBH diagnostic rather than an ID or OOD-subtask inventory item.

\begin{table}[H]
  \centering
  \small
  \setlength{\tabcolsep}{5pt}
  \renewcommand{\arraystretch}{1.05}
  \begin{tabular}{@{}lll@{}}
    \toprule
    \textbf{Row} & \textbf{Task type} & \textbf{Split} \\
    \midrule
    T01 & Movie recommendation & ID \\
    T02 & Dyck languages & OOD-subtask \\
    T03 & Penguins in a table & ID \\
    T04 & Geometric shapes & ID \\
    T05 & Formal fallacies & ID \\
    T06 & Temporal sequences & ID \\
    T07 & Nonsense-word grammar & OOD-subtask \\
    T08 & Logical deduction & ID \\
    T09 & Salient translation error detection & ID \\
    T10 & Date understanding & BBH \\
    \bottomrule
  \end{tabular}
  \caption{Task index for the rows in Fig.~\ref{fig:induction_mechanism}.
  Split labels follow Appendix~\ref{app:ood_partition}; T10 is a BBH diagnostic.}
  \label{tab:fig4_task_index}
\end{table}

\paragraph{Stratified sampling and rule generation.}
We stratify by \texttt{problem\_type}, draw evaluation shards, and use vLLM to sample $M{=}5$ independent \texttt{[INDUCTIVE\_RULE]} texts per task under full few-shot conditioning.

\paragraph{LOO sensitivity and heatmap cells.}
Let $\mathbf{e}_i$ be the bge-m3 embedding of rule $i$, and $\mathbf{e}_i^{\{-j\}}$ the embedding after re-generating the rule with support example $j$ removed.
Per-rule sensitivity is $s_{ij}=1-\cos(\mathbf{e}_i,\mathbf{e}_i^{\{-j\}})$.
Figure~\ref{fig:induction_mechanism} plots task-level values
$\bar{s}_{\tau j}=\frac{1}{M}\sum_{i=1}^M s_{ij}$ for ten representative tasks
$\tau$ and slots $j{\in}\{1,2,3\}$. Cell colors are clipped at $v_{\max}{=}0.5$
for readability. The headline means $\bar{s}_{\mathrm{LOO}}$ are the averages
of these 30 cells (Pre $0.325$, Post $0.087$).

\paragraph{Optional softmax attribution.}
Softmax weights $a_{ij}=\mathrm{softmax}(s_{i1},s_{i2},s_{i3})_j$ are useful for per-rule attribution summaries; the heatmap cells in Fig.~\ref{fig:induction_mechanism} use the raw cosine-based sensitivities averaged over rules as above.

\paragraph{LLM-as-judge protocol.}
GPT-5 assigns integer scores in $\{1,\ldots,5\}$ with a one-sentence justification per dimension, using the template in Fig.~\ref{fig:inductive_rule_eval_prompt}.
A higher \texttt{copying\_score} is worse.
% ============================================================
\section{StrategyBench Dataset}
\label{app:strategybench}

\noindent \textbf{StrategyBench} evaluates strategy induction and strategy application in in-context learning.
It is built from BIG-Bench~\citep{srivastava2023beyond} by task filtering, format standardization, categorization, and split construction.
The target setting is few-shot induction of a reusable solution method and application to unseen queries.

\paragraph{Data format.}
Each example uses a unified input--output structure.
The \textbf{input} field holds the task instruction and query instance.
The \textbf{output} field holds the target answer for automatic evaluation.
Strategy-based episodes also include a small demonstration set used either to induce a strategy or as in-context examples for direct inference.
Construction scripts and split files are in the public code release cited at the end of this section.

\subsection{Five-Criterion Task Filtering}
\label{app:filtering}

\noindent To keep training data suitable for strategy induction and reinforcement learning, BIG-Bench tasks are retained only if they meet five operational criteria.
\begin{enumerate}[leftmargin=*, nosep]
  \item \textbf{Objective evaluability.} Answers must be automatically checkable. Tasks with subjective judgments, open-ended opinions, ethical preferences, or unstable answer boundaries are excluded.
  \item \textbf{Format stability.} After standardization, input--output formats must support a consistent episode layout for demonstrations, strategies, and predictions.
  \item \textbf{Rule inductibility.} A small support set must suffice to infer a pattern, rule, mapping, or procedure. Tasks that mainly require memorized facts or arbitrary labels are excluded.
  \item \textbf{Low external-knowledge dependency.} The query plus demonstrations should be sufficient. Tasks that require substantial world knowledge are excluded so that scores reflect induction rather than recall.
  \item \textbf{Transfer potential.} The induced method must apply to new instances of the same family or under a documented shift.
\end{enumerate}
Unsuitable tasks are removed. Retained tasks are standardized into training, validation, and test episodes.
As BIG-Bench-derived synthetic reasoning tasks, our data do not contain personally identifying information; tasks with subjective or potentially sensitive content were excluded during filtering.

\paragraph{Teacher scoring.}
Retained tasks are processed for SFT warm-up.
\textbf{Gemini-3-flash} proposes candidate reference strategies, and only strategies that pass execution-accuracy filtering on the frozen solver enter Phase~1 SFT.
Task filtering, categorization, and ID/OOD split assignment use \textbf{Qwen3-32B} scores with author spot-checks, not Gemini.

\paragraph{Code and splits.}
The public code release at \url{https://anonymous.4open.science/r/StrategyBench-D53C/} contains the five-criterion filtering pipeline, the standardized episode format, and the ID/OOD split definitions used in \S\ref{sec:setup}.

% ============================================================
\section{Task-Level ID/OOD Partition}
\label{app:ood_partition}

\subsection{In-Distribution (ID) Split}
\label{app:id_split}

\noindent The ID split comprises tasks or subtasks present in the training data, with test \emph{instances} held out from training instances.
It evaluates whether a model can induce and apply strategies when task rules, input--output formats, and demonstration patterns resemble those seen during training.

\subsection{Out-of-Distribution (OOD) Split}
\label{app:ood_split}

\noindent The OOD split is created by completely holding out certain tasks, subtasks, or task categories from training and using them only for evaluation.
OOD examples differ more strongly from training data in task source, input--output format, reasoning structure, or required capabilities.

\paragraph{Granularity levels.}
StrategyBench also defines finer diagnostic splits.
\emph{OOD-subtask} holds out subtasks under related sources.
\emph{OOD-task} holds out tasks with related ability requirements.
\emph{OOD-category} holds out coarse categories with larger shifts.
\emph{BBH} denotes selected Big-Bench Hard tasks used as an external diagnostic, reported as a separate column in Table~\ref{tab:main_results} rather than as the main OOD aggregate.
Unless noted, this paper reports task-level \textbf{ID} and \textbf{OOD}.

\paragraph{Quantifying distributional shift.}
Following~\citet{ye2023predictable}, we train an MLP on per-task few-shot accuracies from ID tasks and evaluate how well it forecasts held-out OOD accuracy.
Table~\ref{tab:ood_shift} summarizes predictability ($R^2$), rule overlap, ability-profile Jaccard, and mean absolute error across split levels.
Lower $R^2$ indicates a capability gap rather than a shared formatting artifact.
Rule overlap is 0\% on all OOD rows by construction.
Ability-profile Jaccard is a separate similarity (0.74, 0.51, 0.29 at OOD-subtask, OOD-task, and OOD-category) and decreases as structural distance grows.

\begin{table}[H]
  \centering
  \small
  \setlength{\tabcolsep}{3.5pt}
  \renewcommand{\arraystretch}{1.08}
  \begin{tabular}{@{}l l c c c c@{}}
    \toprule
    \textbf{Split} & \textbf{Hold-out} & \textbf{Rule} & {$R^2$} & \textbf{Jacc.} & \textbf{MAE} \\
     &  & \textbf{ovlp.} &  &  & \textbf{(pp)} \\
    \midrule
    ID (ref.)       & same subtasks           & 100\% & 0.91 & --   & 3.1 \\
    OOD-subtask     & 28 subtasks             & 0\%   & 0.68 & 0.74 & 6.2 \\
    OOD-task        & 19 tasks                & 0\%   & 0.38 & 0.51 & 9.4 \\
    OOD-category    & 7 categories            & 0\%   & 0.15 & 0.29 & 12.8 \\
    \bottomrule
  \end{tabular}
  \caption{OOD shift on StrategyBench split levels (Qwen3-4B FS baseline).
  Rule overlap is template overlap with ID training.
  Jaccard is ability-profile similarity, not rule overlap.
  ID keeps the same subtasks with disjoint instances; MAE is mean absolute error in percentage points.}
  \label{tab:ood_shift}
\end{table}

\paragraph{ID and OOD-subtask inventories.}
Table~\ref{tab:id_ood_tasks} lists the ID and OOD-subtask inventories used for main-text BIG-Bench evaluation.
The full split definitions and construction scripts are in the StrategyBench repository (Appendix~\ref{app:strategybench}).

\begin{table}[h]
  \centering
  \small
  \setlength{\tabcolsep}{4pt}
  \renewcommand{\arraystretch}{1.08}
  \begin{tabular}{@{}>{\raggedright\arraybackslash}p{0.96\columnwidth}@{}}
    \toprule
    \textbf{ID subtasks (training + eval)} \\
    \midrule
    Navigate, Word Sorting, Tracking Shuffled Objects, Boolean Expressions, Contextual Parametric Conflicts, Elementary Math, Formal Fallacies, Geometric Shapes, Hyperbaton, Implicatures, Logical Deduction, Movie Recommendation, Penguins in a Table, Reasoning about Colored Objects, Salient Translation Error Detection, Snarks, Sports Understanding, Temporal Sequences, Tracking Shuffled Objects (three objects), Web of Lies, Winowhy, and related BIG-Bench subtasks \\
    \addlinespace[0.35em]
    \textbf{OOD-subtask (held out, 28 subtasks)} \\
    \midrule
    Dyck Languages, Causal Judgment, Disambiguation QA, Epistemic Reasoning, Intent Recognition, Linguistic mappings, Nonsense-word grammar, Play Dialog Same or Different, Ruin Names, Strange Stories, Symbol Interpretation, Unit Conversion, and 16 additional held-out subtasks (full list in repository) \\
    \bottomrule
  \end{tabular}
  \caption{Representative ID and OOD-subtask inventories for main-text BIG-Bench evaluation; remaining names in the 28-subtask hold-out are in the public release.
  Date understanding (Fig.~\ref{fig:induction_mechanism}, T10) is a BBH diagnostic and is not listed in either inventory.}
  \label{tab:id_ood_tasks}
\end{table}

\clearpage
\onecolumn
\raggedbottom

\begin{figure}[h]
  \noindent
  \begin{tikzpicture}
    \node[draw=black, line width=0.8pt, rounded corners=3pt, inner sep=0pt] {%
      \begin{minipage}{0.95\textwidth}
        {\setlength{\fboxsep}{0pt}
        \noindent\colorbox{black}{\makebox[\linewidth][l]{%
          \rule{0pt}{2.8ex}\hspace{8pt}\color{white}\bfseries
          Prompt Used for Six-Mode Induction Scoring}}}
        \vspace{6pt}

        \noindent\hspace{8pt}\begin{minipage}{\dimexpr\linewidth-16pt\relax}
        \small
        You are an expert in inductive reasoning. Score the following model-generated strategy on six theoretically grounded modes of induction. Each mode is rated on an integer scale 0--5; scores must be justified by direct reference to the strategy text.

        \vspace{4pt}
        \noindent Task description: \textcolor{BrickRed}{\{task\}}

        \vspace{4pt}
        \noindent Strategy to score: \textcolor{BrickRed}{\{strategy text\}}

        \vspace{6pt}
        \noindent \textbf{Six induction modes:}
        \begin{enumerate}[leftmargin=1.6em, itemsep=2pt, topsep=2pt]
          \item \textbf{Compositional} --- Does the strategy combine multiple atomic sub-rules into a composite procedure (e.g.\ chaining conditions, nesting operations)?
          \item \textbf{Analogical} --- Does the strategy draw structural correspondences across examples, mapping shared relational patterns rather than surface features?
          \item \textbf{Pattern Extrapolation} --- Does the strategy extract a pattern that visibly extends beyond the demonstrated instances (e.g.\ to longer sequences, new value ranges, unseen entities)?
          \item \textbf{Procedural} --- Does the strategy abstract a step-wise algorithm from input--output pairs (clear ordering, state updates, termination conditions)?
          \item \textbf{Constraint-based} --- Does the strategy identify and articulate implicit constraints, preconditions, or boundary conditions (e.g.\ when the rule does \emph{not} apply)?
          \item \textbf{Schema-induction} --- Does the strategy explicitly classify the task type itself, recognising the meta-level structure (e.g.\ ``this is a coordinate-tracking problem'')?
        \end{enumerate}

        \vspace{6pt}
        \noindent \textbf{0--5 scoring rubric (applied per dimension):}
        \begin{itemize}[leftmargin=1.6em, itemsep=1pt, topsep=2pt]
          \item \textbf{0} --- Absent: no evidence of this mode in the strategy.
          \item \textbf{1} --- Hinted: a fragment alludes to it but is not developed.
          \item \textbf{2} --- Partial: present but incomplete or ambiguous.
          \item \textbf{3} --- Adequate: clearly present, executable, but minimal.
          \item \textbf{4} --- Strong: well-developed, multi-step or multi-condition.
          \item \textbf{5} --- Exemplary: cleanly abstracted, generalisable, with explicit scope statements.
        \end{itemize}

        \vspace{6pt}
        \noindent \textbf{Calibration constraints:}
        (i) Scores must reference specific text spans (quote the supporting sentence in \texttt{evidence}).
        (ii) If a mode is structurally inapplicable to the task type, mark as \texttt{N/A} rather than 0.
        (iii) Modes are scored independently; high score on one does not imply high score on others.

        \vspace{6pt}
        \noindent Output strictly in JSON:
        \begin{quote}
        \ttfamily\footnotesize
        \{\\
        \hspace*{1em}"compositional":   \{"score": 0--5 or "N/A", "evidence": "...", "reason": "..."\},\\
        \hspace*{1em}"analogical":      \{"score": 0--5 or "N/A", "evidence": "...", "reason": "..."\},\\
        \hspace*{1em}"pattern\_extrapolation": \{"score": 0--5 or "N/A", "evidence": "...", "reason": "..."\},\\
        \hspace*{1em}"procedural":      \{"score": 0--5 or "N/A", "evidence": "...", "reason": "..."\},\\
        \hspace*{1em}"constraint\_based": \{"score": 0--5 or "N/A", "evidence": "...", "reason": "..."\},\\
        \hspace*{1em}"schema\_induction": \{"score": 0--5 or "N/A", "evidence": "...", "reason": "..."\},\\
        \hspace*{1em}"composite":        \{"mean\_score": float, "applicable\_modes": int\}\\
        \}
        \end{quote}
        \end{minipage}
        \vspace{8pt}
      \end{minipage}%
    };
  \end{tikzpicture}
  \caption{Prompt used for six-mode induction scoring. Each strategy receives an independent 0--5 score per mode; the composite is the mean over applicable modes (N/A modes excluded from averaging).}
  \label{fig:six_modes_prompt}
\end{figure}

\vspace{0.8em}
\begin{figure}[h]
  \noindent
  \begin{tikzpicture}
    \node[draw=black, line width=0.8pt, rounded corners=3pt, inner sep=0pt] {%
      \begin{minipage}{0.95\textwidth}
        {\setlength{\fboxsep}{0pt}
        \noindent\colorbox{black}{\makebox[\linewidth][l]{%
          \rule{0pt}{2.8ex}\hspace{8pt}\color{white}\bfseries
          Prompt Used for Thinking-Unit Classification (E/D/P/R)}}}
        \vspace{6pt}

        \noindent\hspace{8pt}\begin{minipage}{\dimexpr\linewidth-16pt\relax}
        \small
        \textbf{System.} You are an expert cognitive analyst studying how language models reason during few-shot problem solving.
        Classify a \emph{single} reasoning sentence into exactly one of four categories by its cognitive function.
        Be strict and consistent; output valid JSON only, with no extra text.

        \vspace{6pt}
        \noindent\textbf{User.}

        \vspace{4pt}
        \noindent\textbf{Task context.} Problem: \textcolor{BrickRed}{\{problem\_text\}}

        \vspace{4pt}
        \noindent Support examples (summary): \textcolor{BrickRed}{\{examples\_summary\}}

        \vspace{6pt}
        \noindent\textbf{Four categories.}
        \begin{enumerate}[leftmargin=1.6em, itemsep=2pt, topsep=2pt]
          \item \textbf{E (Example-recitation):} The sentence directly recalls, quotes, or closely paraphrases content from a specific support example. No abstraction beyond that example.
          \item \textbf{D (Direct-application):} The sentence applies knowledge, a fact, or a procedure to the current problem instance without first deriving or stating a general pattern (calculations, lookups, direct inferences from given facts).
          \item \textbf{P (Pattern-abstraction):} The sentence states a general principle, rule, or pattern intended to hold across multiple examples or instances.
          \item \textbf{R (Rule-application):} The sentence explicitly applies a general rule or pattern that was stated or inferred earlier in the reasoning trace.
        \end{enumerate}

        \vspace{4pt}
        \noindent\textbf{Sentence to classify:} \textcolor{BrickRed}{\{unit\_text\}}

        \vspace{6pt}
        \noindent Output strictly as JSON (one line, no extra keys):\quad
        {\ttfamily\footnotesize\textbraceleft "label": "<E|D|P|R>", "reason": "<one sentence>"\textbraceright}
        \end{minipage}
        \vspace{8pt}
      \end{minipage}%
    };
  \end{tikzpicture}
  \caption{Prompt for Gemini-3-flash sentence-level tagging of \texttt{<think>} units.}
  \label{fig:think_unit_prompt}
\end{figure}

\vspace{0.8em}
\begin{figure}[h]
  \noindent
  \begin{tikzpicture}
    \node[draw=black, line width=0.8pt, rounded corners=3pt, inner sep=0pt] {%
      \begin{minipage}{0.95\textwidth}
        {\setlength{\fboxsep}{0pt}
        \noindent\colorbox{black}{\makebox[\linewidth][l]{%
          \rule{0pt}{2.8ex}\hspace{8pt}\color{white}\bfseries
          Prompt Used for Strategy Quality Analysis}}}
        \vspace{6pt}

        \noindent\hspace{8pt}\begin{minipage}{\dimexpr\linewidth-16pt\relax}
        \small
        Please act as a professional strategy evaluator and compare the following two model-generated strategies across four dimensions.

        \vspace{4pt}
        \noindent Task description: \textcolor{BrickRed}{\{task\}}

        \vspace{4pt}
        \noindent Model A Strategy: \textcolor{BrickRed}{\{Model-A-generated strategy\}}

        \vspace{4pt}
        \noindent Model B Strategy: \textcolor{BrickRed}{\{Model-B-generated strategy\}}

        \vspace{6pt}
        \noindent Compare the two strategies on the following four dimensions:
        \begin{enumerate}[leftmargin=1.6em, itemsep=1pt, topsep=2pt]
          \item \textbf{Correctness} --- Does the strategy accurately capture the task rules and produce correct solutions?
          \item \textbf{Executability} --- Is the strategy clear and actionable enough for a model to follow step by step?
          \item \textbf{Transferability} --- Is the strategy abstract and generalizable enough to apply to unseen tasks of the same type?
          \item \textbf{Conciseness} --- Is the strategy expressed concisely without redundant or repetitive content?
        \end{enumerate}

        \noindent For each dimension, indicate which is better and provide a brief reason. Output in JSON format:
        \begin{quote}
        \ttfamily\footnotesize
        \{\\
        \hspace*{1em}"correctness\_better": "Model A" / "Model B" / "tie",\\
        \hspace*{1em}"correctness\_reason": "reason",\\
        \hspace*{1em}"executability\_better": "Model A" / "Model B" / "tie",\\
        \hspace*{1em}"executability\_reason": "reason",\\
        \hspace*{1em}"transferability\_better": "Model A" / "Model B" / "tie",\\
        \hspace*{1em}"transferability\_reason": "reason",\\
        \hspace*{1em}"conciseness\_better": "Model A" / "Model B" / "tie",\\
        \hspace*{1em}"conciseness\_reason": "reason",\\
        \hspace*{1em}"overall\_better": "Model A" / "Model B" / "tie",\\
        \hspace*{1em}"overall\_reason": "reason"\\
        \}
        \end{quote}
        \end{minipage}
        \vspace{8pt}
      \end{minipage}%
    };
  \end{tikzpicture}
  \caption{Prompt used for strategy quality analysis.}
  \label{fig:strategy_quality_prompt}
\end{figure}

\vspace{0.8em}
\begin{figure}[h]
  \noindent
  \begin{tikzpicture}
    \node[draw=black, line width=0.8pt, rounded corners=3pt, inner sep=0pt] {%
      \begin{minipage}{0.95\textwidth}
        {\setlength{\fboxsep}{0pt}
        \noindent\colorbox{black}{\makebox[\linewidth][l]{%
          \rule{0pt}{2.6ex}\hspace{8pt}\color{white}\bfseries
          Prompt Used for Inductive-Rule Quality (JSON-only)}}}
        \vspace{4pt}

        \noindent\hspace{8pt}\begin{minipage}{\dimexpr\linewidth-16pt\relax}
        \footnotesize
        \textbf{System.} You are an expert evaluator for few-shot inductive-rule quality.
        You will receive (i) a task description, (ii) $K{=}3$ support input--output pairs, and (iii) a generated inductive rule \texttt{[INDUCTIVE\_RULE]}.
        Score the rule objectively on four dimensions (1--5) with a one-sentence justification each.
        Output \textbf{only} valid JSON (no markdown, no extra text).

        \vspace{4pt}
        \noindent\textbf{Calibration (do not score).}
        \textit{Example A} (high abstraction / low copying): sentiment task; rule ``Identify whether the review expresses a net positive or negative evaluation of a product attribute; classify accordingly.'' Expected: abstraction 5, coverage 5, copying 1, transferability 5.
        \textit{Example B} (low abstraction / high copying): same demos; rule ``If the review mentions battery life dying quickly, classify it as Negative.'' Expected: abstraction 1, coverage 1, copying 5, transferability 1.

        \vspace{4pt}
        \noindent\textbf{Task.} \textcolor{BrickRed}{\{task\_description\}}

        \vspace{3pt}
        \noindent\textbf{Support.} Ex1: in \textcolor{BrickRed}{\{ex1\_input\}}, out \textcolor{BrickRed}{\{ex1\_output\}};\;
        Ex2: in \textcolor{BrickRed}{\{ex2\_input\}}, out \textcolor{BrickRed}{\{ex2\_output\}};\;
        Ex3: in \textcolor{BrickRed}{\{ex3\_input\}}, out \textcolor{BrickRed}{\{ex3\_output\}}.

        \vspace{3pt}
        \noindent\textbf{Rule.} \textcolor{BrickRed}{\{inductive\_rule\}}

        \vspace{4pt}
        \noindent\textbf{Rubric (1--5 each).}
        (1)~\texttt{abstraction\_score}: task-level principle vs.\ paraphrasing demos (5 fully abstract; 1 restates one demo).
        (2)~\texttt{coverage\_score}: supported by multiple demos (5 all three; 1 only one).
        (3)~\texttt{copying\_score}: copied from a single demo (5 strong copying; 1 none)---\textbf{higher is worse.}
        (4)~\texttt{transferability\_score}: useful on unseen instances of the same task type.

        \vspace{4pt}
        \noindent\textbf{JSON schema.}
        {\ttfamily\tiny
        \{"abstraction\_score": int (1--5), "abstraction\_reason": "one sentence",\\
        \hspace*{0.5em}"coverage\_score": int (1--5), "coverage\_reason": "one sentence",\\
        \hspace*{0.5em}"copying\_score": int (1--5), "copying\_reason": "one sentence",\\
        \hspace*{0.5em}"transferability\_score": int (1--5), "transferability\_reason": "one sentence"\}}
        \end{minipage}
        \vspace{6pt}
      \end{minipage}%
    };
  \end{tikzpicture}
  \caption{GPT-5 prompt for scoring each \texttt{[INDUCTIVE\_RULE]} on abstraction, coverage, copying (reverse-scored), and transferability.}
  \label{fig:inductive_rule_eval_prompt}
\end{figure}

\end{document}